\documentclass[runningheads]{llncs}
\usepackage{graphicx}
\usepackage{algpseudocode}
\usepackage{algorithm}
\usepackage{mathtools}
\usepackage{xcolor}

\usepackage{amsmath}
\usepackage{amsthm}
\usepackage{amsfonts}
\usepackage{amssymb}
\usepackage{appendix}
\usepackage{multirow}
\usepackage{threeparttable}
\usepackage{microtype}

\usepackage{xcolor,colortbl}
\definecolor{light-gray}{gray}{0.7}

\newcommand{\dnote}[1]{}
\newcommand{\anote}[1]{}
\newcommand{\mnote}[1]{}

\makeatletter
\newcommand{\printfnsymbol}[1]{%
  \textsuperscript{\@fnsymbol{#1}}%
}
\makeatother

\begin{document}
\title{Transparency Tools for Fairness in AI (Luskin\thanks{Our suite of tools
 is christened ``Luskin,'' as
 this project began at the UCLA Meyer and Renee Luskin Conference Center.})}
\titlerunning{Transparency Tools for Fairness in AI (Luskin)}
\author{
Mingliang Chen\inst{2}\thanks{These authors contributed equally to this work.}
%\orcidID{1111-2222-3333-4444}
\and
Aria Shahverdi\inst{2}\printfnsymbol{2}
\and
Sarah Anderson\inst{1}
%\orcidID{0000-1111-2222-3333}
\and
Se Yong Park\inst{2,4}
\and
Justin Zhang\inst{2,4}
\and
Dana Dachman-Soled\inst{2}
\thanks{Dana Dachman-Soled and Aria Shahverdi were supported in part by NSF grants \#CNS-1933033,  \#CNS-1840893, \#CNS-1453045 (CAREER), by a research partnership award from Cisco and by financial assistance award 70NANB15H328 from the U.S. Department of Commerce, National Institute of Standards and Technology.}
\and
Kristin Lauter\inst{3}
\and
Min Wu\inst{2}
}
%Third Author\inst{3}\orcidID{2222--3333-4444-5555}}
%
\authorrunning{M.~Chen et al.}
\institute{
University of St.~Thomas,
St.~Paul, MN, USA
%\email{}
\and
University of Maryland,
College Park, MD, USA
%\url{http://www.springer.com/gp/computer-science/lncs} 
\and
Microsoft Research,
Redmond WA, USA\\
\and
Montgomery Blair High School,
Silver Spring, MD, USA
}

\maketitle              % typeset the header of the contribution
\begin{abstract}
We propose new tools for policy-makers to use when assessing and correcting fairness and bias in AI algorithms.  The three tools are:

\vspace{.8mm}

\begin{itemize}
    \item A new definition of fairness 
    called ``controlled fairness''
    with respect to choices of protected features and filters.
    The definition provides a simple test
    of fairness of an algorithm 
    with respect to a dataset.
    This notion of fairness is suitable
    in cases where fairness is prioritized
    over accuracy, such as in cases where there is no ``ground truth'' data, only
    data labeled with past decisions (which may have been biased).
    \item Algorithms for retraining a given classifier to achieve ``controlled fairness'' with respect to a choice of features and filters. Two algorithms are presented, implemented and tested.
    These algorithms require training two different models in two stages. We experiment with combinations of various types of models for the first and second stage and report on which combinations
    perform best in terms of fairness and accuracy.
    \item Algorithms for adjusting model parameters to achieve a notion of fairness called ``classification parity.''
    This notion of fairness is suitable in cases where accuracy is prioritized. Two algorithms are presented, one which assumes
    that protected features are accessible to the model during testing,
    and one which assumes protected features are not accessible
    during testing.
\end{itemize}

\vspace{.8mm}

We evaluate our tools on three different publicly available
datasets.
We find that the tools are useful for understanding various dimensions of bias, and that in practice the algorithms are effective in starkly
reducing
a given observed bias when tested on new data.

\keywords{Controlled Fairness \and Machine Learning \and Policy.}
\end{abstract}

\section{Introduction}
Machine Learning (ML) is a 
set of
valuable mathematical 
and algorithmic
tools, which use existing data (known as ``training data'') to learn a pattern, and predict future outcomes on new data based on that pattern.  ML algorithms are the foundation of a revolution in Artificial Intelligence (AI), which is replacing humans with machines.
In business, public policy and health care, decision-makers  increasingly rely on the output of a trained ML classifier to make important and life-altering decisions such as whether to grant an individual a loan, parole, or admission to a college or a hospital for treatment.
In recent years it has become clear that, as we use AI or ML algorithms to make predictions and decisions, we perpetuate bias that is inherent in the training data.

In this paper, we propose new tools for policy-makers to use when assessing and correcting fairness and bias in AI algorithms.  Datasets typically come in the form of databases with rows corresponding to individual people or events, and columns corresponding to the features or attributes of the person or event.  Some features are considered sensitive and may be ``protected,'' such as race, gender, or age.  Typically we are interested in preventing discrimination or bias based on ``protected'' features, at least in part due to the fact that it is illegal.  Other features may be either highly correlated with protected features or may be relevant from a common sense point of view with the decision to be made by the classifier.
By ``protected classes''
we refer to the different groups
arising from 
the various settings of
the protected feature.

The notion of fairness is a rather complex one and there are multiple aspects of fairness and/or perceived fairness related to machine learning (cf.~\cite{kleinberg2016inherent,NIPS:JKMR16,hardt2016equality,KDD:FFMSV15,ITCS:DHPRZ12}). 
This work is not intended as a survey 
and therefore we discuss only the 
fairness definitions most closely related
to the current work.
We focus on two angles, and argue that
each is appropriate in different real-life settings:

\paragraph{Controlled Fairness.}
In cases where a machine learning algorithm is trained on data labeled with prior
\emph{decisions}, as opposed
to an objective ``ground truth,''
our main concern is \emph{parity},
across protected classes.
For example, in a stop-question-frisk law enforcement setting,
a classifier deciding whether or not a person should be frisked,
is trained on past stop-question-frisk data.
But whether a person was frisked or not in the past, depends on a (potentially biased) decision.
Thus, this is a setting where
achieving accuracy with respect to past decisions may not be desirable.
Another setting in which \emph{parity}
is our main concern (taking precedence over accuracy) is a setting in which an organization
is compelled to bear the risk of an incorrect decision,
in order to improve societal welfare.
For example, a bank may use a classifier to predict whether a person
will default on a loan, and then use this information to determine whether
to approve the loan.
In this case, our main concern may be that the classifier achieves
\emph{parity} across protected classes,
as long as the bank's overall risk does not significantly increase.

Our fairness definition
requires that
controlling for certain ``unprotected attributes'' such as
\emph{type of crime} in the law-enforcement example, or
\emph{education} level in the loan example, the classifier's $0/1$
output rate with respect to a given dataset is approximately
the same across the protected classes.
We believe this definition will be useful to practitioners
as it provides an explicit test for fairness of
a specific algorithm with respect to a specific dataset.
Of course, we still want to achieve the best accuracy possible with respect to the prior decisions
(since the
point of training the machine learning model in the first place is to obtain
an algorithm that emulates human
decisions), while ensuring that the fairness conditions are met.

By controlling for ``unprotected features,''
we are taking the point of view that it can make sense to ``filter'' datasets based on ``unprotected features'' which seem relevant to making a good decision. 
Filtering simply selects various rows which satisfy certain conditions on the entries in specified columns.  This can be viewed as a controlled experiment, which allows one to determine
the bias
stemming from the protected feature, as opposed to other confounding factors.
After filtering, we propose to test for fairness with respect to protected features by checking whether ratios of outcomes are approximately the same across the protected classes.

\paragraph{Accuracy-Based Fairness.}
In cases where accuracy with respect to a ground truth is prioritized, the above definition may not be the right choice. In this case, the notion of fairness we consider ensures that the true positive rate (TPR) and false positive rate (FPR) are as close as possible across the protected classes, while overall accuracy remains high. We enforce this equivalent learning performance across all protected classes to the extent possible, sacrificing little on the high performance of the model on the majority protected class. This coincides with the \emph{equalized odds}~\cite{hardt2016equality} and \emph{classification parity}~\cite{corbett2018measure} notion that have previously been considered in the literature. \mnote{In this case, the notion of fairness we consider ensures that the true positive rate (TPR) and false positive rate (FPR) are as close as possible across the protected classes, while overall accuracy remains high. This coincides with the \emph{classification parity} notion
that has previously been considered in the literature~\cite{corbett2018measure}.}
An appropriate setting for applying this notion of fairness is
in predicting recidivism, where there is arguably a more objective ``ground truth.'' Specifically, taking as an example the parole decisions,
the machine learning algorithm is not trained on data labeled by the decisions themselves, but rather on data labeled according to whether or not a person who was released was subsequently re-arrested.
Furthermore, the risk of making an incorrect decision is extremely high and cannot simply be absorbed as a loss, as in the case of loan defaults.
It has been documented (and our own experiments support this) that in the recidivism case, when machine learning algorithms are trained on available data, non-whites have both higher TPR and FPR than whites. This implies that a larger fraction of non-whites than whites are predicted to re-commit a crime, when, in fact, they do not go on to re-commit a crime within two years. Equalizing the TPR and FPR for whites and non-whites (to the extent possible) --- and simultaneously maintaining the overall prediction accuracy of the classifier --- results in improved fairness outcomes across the protected classes.

\newpage

\subsection{The Toolkit}
The three tools we introduce in this work are:
\begin{itemize}
    \item A new definition of \emph{controlled fairness}
    with respect to choices of protected features and filters. This definition provides a simple
    test of fairness of an algorithm with respect
    to a dataset.
    \item Algorithms for retraining a classifier to achieve ``controlled fairness'' with respect to a choice of features and filters. Two algorithms are presented, implemented and tested.
    These algorithms require training two different models in two stages. We experiment with combinations of various types of models for the first and second stage and report on which combinations
    perform best in terms of fairness and accuracy.
    \item Algorithms for adjusting model parameters to achieve ``classification parity.''
    Two algorithms are presented, one which assumes
    that protected features are accessible to the model during testing,
    and one which assumes protected features are not accessible during testing.
\end{itemize}

We also implement and evaluate our tools on the Stop-Question-Frisk dataset from NYPD, the Adult Income dataset
from the US Census and the COMPAS recidivism dataset.
The Stop, Question and Frisk dataset is a public record of an individual who has been stopped by NYPD, and it contains detailed information about the incident, such as time of the stop, location of the stop, etc. The Adult Income dataset was taken from the 1994 Census Database, and each row has information about an individual such as marital status, education, etc. The COMPAS recidivism dataset records individuals' basic information and their recidivism within two years. The basic information includes race, age, crime history, etc.

We find that the tools are useful for understanding various dimensions of bias, and that in practice the tools are effective for eliminating a given observed bias when tested on new data.

\iffalse
\dnote{This appears later.}
\dnote{Fill in the following:}
Main difference in our approach: we work directly with concrete datasets and concrete classifiers to put these definitions into a context where a policy maker can view the implications of filtering on various attributes.
\fi

\subsection{Prior Fairness Definitions and Our Contributions}

%How does one define fairness, especially in settings when the dataset itself is known to exhibit bias due to historical discriminatory practices or for other reasons? 

%
%\textbf{Placeholder: Need to provide more background including recent survey articles: \texttt{https://fairmlclass.github.io/}.}
Establishing a formal definition of fairness
that captures our intuitive notions and ensures
desirable outcomes in practice,
is itself a difficult research problem.
The legal and machine learning literature has proposed different and conflicting definitions of fairness~\cite{kleinberg2016inherent,NIPS:JKMR16,hardt2016equality,KDD:FFMSV15,ITCS:DHPRZ12}.
We begin with an
overview of some definitions of fairness from the literature and their relation to the notions studied in this work.

We first discuss the relationship of our new notion of
``controlled fairness'' to the prior notion of
``statistical parity,'' originally introduced in~\cite{ITCS:DHPRZ12}.
We then discuss our contributions with respect to definitions, algorithms and implementations in this regime.

Next, we discuss the prior notion of ``classification parity''~\cite{corbett2018measure} and its relationship to our work on accuracy-based fairness.
We discuss our contributions with respect to definitions, algorithms and implementations in this regime.

\iffalse
\dnote{check notation.}
We will assume features can be categorized into protected and unprotected features
%\footnote{We also note that features in the unprotected class may be related to features in the protected class which raises additional concerns. We make a critical assumption for now that all features are independent of one another.} 
denoted as $x=(x_p, x_u)$ for protected and unprotected respectively. 
\fi

\subsubsection{Statistical parity.}
The notion of
statistical parity stems from
the legal notion of ``disparate impact''~\cite{barocas2014datas}.
Other names in the literature for the same
or similar notion
include \emph{demographic parity}~\cite{ICDM:CalKamPec09,arXiv:Zliobaite15}
and \emph{group fairness}~\cite{ICML:ZWSPD13}.
This notion essentially requires that
the outcome of a classifier is
equalized across the protected classes.
For example, it may require that the percentage of female
and male applicants accepted to a college is approximately the same.

More formally, the \emph{independence} notion that
underlies the definition of statistical parity
(see e.g.~\cite{TB:BarHarNar19})
requires that,
in the case of binary classification:
\begin{equation} \label{eq:indep}
\mathbb{P}\{R=1\mid A=a\}=\mathbb{P}\{R=1\mid A=b\}.
\end{equation}
Here, $A$ corresponds to the protected feature,
which can be set to value $a$ or $b$.
$R$ is the random variable corresponding to the output
of the classifier on an example sampled from some
distribution $\mathcal{D}$.
Thus, the left side of Equation (\ref{eq:indep}) corresponds to the
probability that a classifier outputs $1$
on an example sampled from $\mathcal{D}$,
\emph{conditioned on the protected feature of the example being set to value $a$}.
Similarly, the right right side of
Equation (\ref{eq:indep}) corresponds to the
probability that a classifier outputs $1$
on an example sampled from $\mathcal{D}$,
\emph{conditioned on the protected feature of the example being set to value $b$}.

Assuming rows of a database are sampled as
i.i.d.~random variables from distribution $\mathcal{D}$,
the left hand side probability
in (\ref{eq:indep})
can be approximated as the \textbf{ratio of}
\emph{the number of rows in the database
for which the classifier outputs $1$ and the protected
feature is set to $a$} \textbf{to}
\emph{the number of rows
in the database for which the protected feature is set to $a$}.
Analogously, the right hand side probability
in (\ref{eq:indep})
can be approximated as the \textbf{ratio of}
\emph{the number of rows
for which the classifier outputs $1$ and the protected
feature is set to $b$} \textbf{to}
\emph{the number of rows for which the protected feature is set to $b$}.
These empirical ratios are the basis of our
controlled fairness definition (which will be introduced formally in Section~\ref{sec:control}).

The independence-based notions discussed above have been criticized in the machine learning literature, mainly due to the fact that
they inherently sacrifice accuracy, since
the \emph{true classification} may itself be deemed
``unfair'' under these definitions.
Our notion of ``controlled fairness''
is a refinement of these notions (as it allows, in addition,
filtering on unprotected features), and 
indeed may preclude achieving perfect accuracy.
Due to this limitation, our notion should only be applied in situations where ``fairness'' is prioritized over ``accuracy.''
For example, in some situations such as college 
admissions, there is no ``ground truth'' to measure accuracy against, only data
about previous admission decisions, which may themselves
have been biased.
Therefore, in such settings,
our goal should not be solely to achieve
optimal accuracy with respect to past decisions.

Other objections to these notions 
include the fact that
it may not always be desirable
to equalize the outcomes across the protected classes.
In our above example,
if 20\% of the female applicants
have GPA of at least 3.5 and SAT scores of at least 1500,
while only 10\% of the male applicants
have GPA of at least 3.5 and SAT scores of at least 1500,
then one may argue that it is ``fair''
for 
a larger percentage of the female applicants 
to be accepted than the male.

%The three main fairness definitions, as outlined in \cite{corbett2018measure}, include:
% To-Do: Discuss broader literature: \cite{kleinberg2016inherent}, Dwork paper, Intersectional Fairness arxiv paper, RAND report, and others
\iffalse
\begin{enumerate}
    \item Anti-classification: assigns predictions without considering protected features,
    \item Classification Parity: assigns predictions that equalizes a pre-determined misclassification error (e.g. false positive rates),
    \item Calibration: assigns predictions such that outcomes are independent of protected features, conditional on a prediction. 
\end{enumerate}

We also acknowledge that bias can arise in different forms \textbf{[placeholder, check for larger literature on causes of biased data]}. With a growing recognition of the need to develop predictive models that address the challenges of bias, this paper's aim is to propose a formal definition of differential fairness, motivated in part by differential privacy \cite{}. Unlike previous definitions, 
\fi

Our notion of fairness remedies exactly this situation,
by allowing ``controls'' or ``filters'' to be placed on
features that are considered ``unprotected.''
In the above example, a ``filter'' selects the set of rows
satisfying the condition that
the GPA is at least 3.5 and SAT score at least 1500.
Then, among those selected rows, we require that the
percentage of accepted females and males is approximately the same.

In summary,
our work on \emph{controlled fairness} makes the following contributions and/or distinctions beyond the notion of
statistical parity and other notions previously
considered in the literature:
\begin{enumerate}
    \item We introduce a definition of ``controlled fairness'' that allows one to identify a ``filtering'' condition on unprotected features that permits one to enforce parity of outcomes
    across subgroups of the protected classes. (Unlike other definitions in \cite{foulds2018intersectional} which enforces an intersection on protected features.)
    This addresses some of the objections to the
    statistical parity notion, since it
    does not mandate na\"{i}ve equalization across the protected classes, but allows for subtleties in how fairness across protected classes is evaluated.
    \item Our proposed notion prioritizes equality in predictions, and does not consider a data-generating mechanism or a ground-truth data. We emphasize that the focus is on stating whether a classifier achieves the controlled fairness notion with respect to a given dataset. Thus, we are not mainly concerned with accuracy
    in this setting and consider it especially appropriate
    to apply this notion in cases where there is no 
    ``ground truth'' data, but only data on prior decisions
    (which may have been biased).
    \item We propose two algorithms to
    retrain a classifier in settings when the originally trained classifier is not fair
    with respect to our fairness notion.
    \item We test our algorithms on real-world data
    and report the outcomes.
    Our algorithms require training two different models in two stages. We experiment with combinations of various types of models for the first and second stage and report on which combinations
    perform best in terms of fairness and accuracy.
\end{enumerate}

\paragraph{Comparison with $\epsilon$-conditional parity.}
We note that our proposed definition is perhaps
most similar to the notion of
\emph{$\epsilon$-conditional parity},
introduced by~\cite{ritov2017conditional}.
The main difference between our notions is that our
notion is concrete: It specifies whether a specific
classifier does or does not achieve 
``controlled fairness'' with respect to a specific dataset.
On the other hand, the notion of~\cite{ritov2017conditional} defines the fairness of a classifier with respect to its output distribution on instances
drawn from various
conditional distributions.
This makes it less useful as a tool
for checking and enforcing fairness with respect to a particular
dataset.

\subsubsection{Classification parity.}
Other notions of fairness considered in the machine learning
literature include
\emph{classification parity,
calibration, calibration within groups} and
\emph{balance for the positive/negative class}~\cite{corbett2018measure,kleinberg2016inherent}.
These notions of fairness prioritize accuracy with respect to a ``ground truth,'' since a classifier that outputs the true labels will always satisfy
the definition.
In our work, we recognize the subtleties in the application of the various fairness notions
(i.e.~not every fairness notion is suitable
for application in all situations).
We therefore consider applying these accuracy-based notions,
specifically, 
the notion of \emph{classification parity}
(as opposed to our previously introduced notion
of controlled fairness),
in the case
that accuracy is prioritized.
\emph{Classification parity}
ensures that 
certain common measures of predictive performance are (approximately) equal across the protected groups. Under this definition, a classifier that predicts recidivism, for example,
would
be required to produce similar false positive rates for white and black parole applicants.
In particular, we will focus on obtaining a classifier whose learning performances across the protected classes are as similar as possible. In other words, we should not be able to distinguish which class of instances is being tested just from the learning performance statistics.
\iffalse
\mnote{The original text: In particular, we will focus on 
obtaining classifiers whose ROC curves are as similar as possible
for the protected and unprotected class.
In other words, for any fixed true positive rate,
we want to ensure that the corresponding false positive rate for the two
classes is as close as possible.}
\fi

In summary, our work on \emph{classification parity} has the following contributions:
\begin{enumerate}
    \item We formulate a \emph{weak} and \emph{strong} condition on classification parity and propose two approaches for achieving improved fairness in classifiers.
    \item The first algorithm trains a 
    single classifier and then tunes the decision thresholds for each of the protected classes to achieve classification parity.
    This methodology can be shown to improve
    fairness for each of the protected classes by equalizing the true positive/false positive rates across classes.
    However, in order to know which thresholds to apply on an input
    instance, the classifier must know the values of the protected
    features.
    \item The second algorithm is applicable in the case that it is not
    legally or socially acceptable to use different classifiers for each of the protected classes, or in the case in which the protected features are simply not known to the classifier.
    In the algorithm, a single classifier is trained by incorporating a trade-off between accuracy and fairness. The fairness is quantitatively measured by the \emph{equalized distribution} (a notion we introduce in Section~\ref{sec:classification_parity}) of positive/negative instances across the protected classes. 
    \iffalse
    \mnote{The original text: In this case, the learning algorithm is modified so that the loss function incorporates not only a measure of the accuracy of the classifier (as is typically the case), but also a ``fairness measure'' that provides a quantitative score of fairness for a given risk assignment algorithm. To quantitatively measure fairness, we use a bucketing approach to partition the range of possible risk scores
    among positive (resp.~negative) instances.
    Two vectors are constructed for the positive (resp.~negative) instances. Each position of the first vector corresponds to the percentage of positive (resp.~negative) instances
    within the protected class that are assigned a score within the corresponding bucket, while each position of the second vector corresponds to the percentage of positive (resp.~negative) instances
    within the unprotected class that are assigned a score within the corresponding bucket. 
    Fairness is quantified by computing the $L_2$ distance between the pairs of vectors and adding the distances. The more ``fair'' a classifier is, the lower the sum of the distances. We note that
    this algorithm also allows for ``tuning''
    of the parameters of the loss function 
    to specify the preference given to accuracy versus fairness of the final classifier.}
    \fi
\end{enumerate}

\subsection{Additional Related Work}

\paragraph{Prior work on achieving statistical parity
and removing discrimination.}
%In this work we present two algorithms for retraining classifiers
%to achieve our notion of controlled fairness, which is closely related to the notion of statistical parity (as discussed above).
Prior work has suggested entirely different algorithms
to achieve similar goals as the goal of this work, which is to obtain classifiers that
achieve the controlled fairness definition.
For example,
the recent work of Wang et al.~\cite{ICML:WanUstCal19},
focuses on the statistical parity notion and
suggests to search for a perturbed distribution,
which they call a ``counterfactual distribution'' on which
disparity of the classifier across the classes is minimized.
Then, for each input to the classifier, they perform a pre-processing step that modifies the features of the input
example according to the counterfactual distribution and then
run the original classifier on the modified input.
The recent work of Udeshi et al.~\cite{ASE:UdeAroCha18} takes as input
a potentially unfair classifier and searches the input
space to find ``discrimatory examples''---two inputs that are highly similar, differ with respect to the protected feature, and are classified differently.
Then, using ``corrected'' labels on these discriminatory examples,
the original classifier can be retrained to improve its
performance.

In contrast to the work of Wang et al.~\cite{ICML:WanUstCal19}, our approach does not require a preprocessing step to be applied to the test input by the end-user. Instead, the final classifier can run as before on a test input. This allows for simplicity and backwards compatibility for the end-user, and would be a more socially acceptable solution, since no overt modification of inputs is performed by the end-user (the only modifications occur during training).

In contrast to both the works of Udeshi et al.~\cite{ASE:UdeAroCha18} and Wang et al.~\cite{ICML:WanUstCal19}, our approach is conceptually simple, and can be performed by running a standard training algorithm as-is to generate the final model that is outputted. There is no additional search step that receives the description of the model and must find either the so-called ``counterfactual distribution,'' or 
``discriminatory examples,'' both of which require additional complex and non-standard algorithms. This makes our approach more suitable as part of a toolkit for policy-makers.

\paragraph{Prior work on achieving accuracy-based fairness.}
A large number of researches have investigated different variations of accuracy-based fairness and improved the fairness achieved by classifiers. These fair algorithms mainly fall into two categories: those that have prior knowledge of the protected feature in test stage~\cite{hardt2016equality,dwork2018decoupled} and those that do not have prior knowldge~\cite{zemel2013learning,calmon2017optimized,zafar2017demographic,zafar2017fairness}. 

Hardt et al.~\cite{hardt2016equality} focused on tuning the classifier to satisfy fairness constraints after the classifier has been trained. The basic idea is to find proper thresholds on receiver operating characteristic (ROC) curve where the classifier meets classification parity. Dwork et al.~\cite{dwork2018decoupled} proposed an algorithm to select a set of classifiers out of a larger set of classifiers according to a joint loss function that balances accuracy and fairness. The advantage is that two works can be applied to any given classifiers since no retraining or modification is needed for the classifiers. However, these methods require access to the protected feature in test stage. Hardt et al.~\cite{hardt2016equality} did not balance the accuracy and fairness during the process of threshold searching, and also does not guarantee to obtain the 
%one and only one solution. 
optimal solution.
Dwork et al.~\cite{dwork2018decoupled} needed large overhead in training multiple classifiers for all the data groups. In contrast, our classification parity based method only requires one trained classifier and the post-processing can provide the unique optimal solution balancing accuracy and fairness.

Learning a new representation for the data is another approach in fairness learning~\cite{zemel2013learning,calmon2017optimized}, removing the information correlated to the protected feature and preserving the information of data as much as possible. For example, Zemel et al.~\cite{zemel2013learning} introduced a mutual information concept in learning the fair representation of the data. The advantage of this approach is that it can be applied before the classifier learning stage and no prior knowledge of the protected feature is required during test-time. The weakness is the lack of the flexibility in accuracy and fairness tradeoff.

A direct way to tradeoff accuracy and fairness is to incorporate a constraint or a regularization of fairness term into the training optimization objective of the classifier. The fairness term is described as classification parity, such as demographic parity~\cite{zafar2017demographic}, equalized odds~\cite{hardt2016equality} and predictive rate parity~\cite{zafar2017fairness}. Note that these methods also do not require access to the protected feature in the test stage. In our work, we propose a new definition of fairness that ensures a strong version of classification parity, which we call \emph{equalized distribution}. Compared with the prior art on classification parity, such as equalized odds~\cite{hardt2016equality}, equalized distribution is a stronger condition on classification parity, requiring that the classifier achieves equivalent performance statistics (e.g., TPR and FPR) among all the groups, independent of the decision threshold settings of the classifier.

%\dnote{Mingliang's summary:
%1. The paper’s method need the prior knowledge of the groups. The final classifier is a set of classifiers for each group of the samples.

%2. The paper’s method first trains multiple different classifiers in each group of the sample (the difference refers to the number of positive classification results). Then among all the combination of the classifiers, the paper find the best one set according to the joint loss function.

%3. The joint loss function has two terms: accuracy term and fairness term. The fairness term the paper proposed is the equality of the number of positive classification across groups.\\

%===============================\\

%Our method with prior knowledge of the groups

%1. We only train one model for the whole dataset, but assign different decision threshold  for different groups. The training overhead is much less than the paper’s method.

%2. Our method regularizes on the equality of TPR and FPR across the groups, which, in my opinion, is closer to “real” fairness.

%}

\subsection{Organization}
The rest of this paper is organized as follows. In Section~\ref{sec:prelim} we introduce the notation we use throughout the paper. In Section~\ref{sec:control}, we begin by presenting our new definition of \emph{controlled fairness}. 
We then present two algorithms to obtain classifiers that achieve the controlled fairness notion
(See Sections~\ref{sec:alg1} and \ref{sec:alg2}).
Next, we 
describe the datasets used to evaluate the performance of our algorithms in Section~\ref{sec:dataset}.
In Section~\ref{sec:cl_choices} we discuss the reasoning behind our high-level implementation choices
and in Section~\ref{sec:exp} we evaluate the performance---fairness and accuracy---of the proposed algorithms. In Section~\ref{sec:classification_parity} we propose two algorithms to achieve classification parity in classifier training, and we show the experimental results on the fairness improvements with the proposed methods.

\section{Preliminaries}\label{sec:prelim}

\paragraph{Notation.}
Dataset $D$ is represented as a matrix in
$\mathbb{R}^{s \times n}$.
It has features $\{f_1, \ldots, f_n\}$ (corresponding
to columns $1, \ldots, n$),
rows $D^1, \ldots, D^s$ and columns
$D_1, \ldots, D_n$.
The $j$-th entry of the $i$-th row is denoted by $D[i][j]$.
We denote by $\#\{D\}$ the number of rows in $D$
(similarly for a set $\mathcal{S}$, we denote by
$\# \mathcal{S}$ the cardinality of the set).
In practice, some features are considered protected,
while others are considered unprotected.
We denote by $F^P \subseteq \{f_1, \ldots, f_n \}$ the set of protected features
and by $F^N \subseteq \{f_1, \ldots, f_n \}$
the set of unprotected features.

\paragraph{Database operators.}
In the database literature, 
the output of a SQL ``Select'' query with condition
$cond$ on dataset $D$ is represented as the output of
operator $\sigma_{cond}(D)$.
A SQL ``Select'' query with condition
$cond$ on dataset $D$ returns a dataset that consists of
all rows of $D$
satisfying condition $cond$, i.e.~if
the entries in certain columns satisfy the
condition set for those columns.
Thus, $|\sigma_{cond}(D)|$ means the number of rows of $D$
satisfying condition $cond$.

%\dnote{TO DO: define $\sigma$ operator.}

\paragraph{Classifiers and Risk Assignment Algorithms}
A Classifier is an algorithm which takes as input a row in dataset and returns a discrete set of values as its output (in our case we assume binary classifer--yes/no), representing the predicted class for the input. Similarly, a risk assignment algorithm takes a input a row in dataset and returns a real number (interpreted as a probability of being a yes/no
instance) between
$0$ and $1$. A risk assignment algorithm is denoted by $A$ and the corresponding classifier is represented by $A^{C}$. Specifically, classifier $A^{C}$ is obtained from risk assignment algorithm $A$ as follows,

\[
    A^C(D^i)= 
\begin{cases}
    1,& A(D^i)   \geq \mathsf{threshold}\\
    0,              & \text{otherwise}
\end{cases}
\]

\iffalse
We sometimes refer to $A$ as a risk assignment algorithm corresponding to $A^{C}$, what this really means is that classifier $A^{C}$ was obtained from risk assignment algorithm $A$.
\fi

\iffalse
\dnote{Classifier outputs one of a discrete set of outputs
(in our case we assume binary classifer--yes/no) whereas 
a risk assignment algorithm outputs a real number between
$0$ and $1$, interpreted as a probability of being a yes/no
instance.}
\fi

\paragraph{Syntax for learning algorithms.}
A learning algorithm $\mathcal{M}$
is an algorithm that takes as input a 
\emph{labeled dataset} $D^+ \in \mathbb{R}^{s \times (n+1)}$
and outputs a binary classifier $C$.
The labeled dataset $D^+$ consists of 
rows $D^{1,+}, \ldots, D^{s,+}$ and
columns
$D^+_1, \ldots, D^+_{n+1}$.
Each row $D^{i,+} \in \mathbb{R}^n \times \{0,1\}$
is called an \emph{example}.
An example $D^{i,+} = D^i || b^i$ consists of
a feature vector $D^i \in \mathbb{R}^n$ that corresponds to a setting of the features
$\{f_1, \ldots, f_n\}$ and a label
$b^i$.\\

\section{Controlled Fairness}\label{sec:control}
In this section, we introduce our new fairness definition,
present algorithms for obtaining classifiers that achieve this notion and evaluate the performance of our algorithms on real datasets.

%\subsection{New Definition: Controlled Fairness}\label{sec:def}

We introduce the following definition of controlled fairness for a binary classification setting:

\iffalse
%\footnote{We assume here that all features are independent of one another.}. 
\begin{definition} [Controlled Fairness]
Let $D \in \mathbb{R}^{s \times n}$ be a dataset,
%with
%features $\{f_1, \ldots, f_n\}$.
let
$cond_N$ (resp.~$cond_P$) be a set of conditions on unprotected features $F^N$
(resp.~protected features $F^P$).
Let $D^+ \in \mathbb{R}^{s \times (n+1)}$
denote the \emph{labeled} dataset obtained
by running (binary) classifier $C$ on dataset $D$.

Define the 
datasets $\mathsf{DP}, \mathsf{DN}$
output by SQL ``Select'' queries as $\mathsf{DP} := \sigma_{cond_P, cond_N}(D^+)$
\anote{is this typo? should it  be $\mathsf{DP} := \sigma_{cond_P}(D^+)$}
and
$\mathsf{DN} := \sigma_{cond_N}(D^+)$.
We require that
$\#\{\mathsf{DP}\}, \#\{\mathsf{DN}\} > 0$.

We say that
C is an $(\epsilon, \delta)$ differentially fair classifier with respect to $D$, $cond_P$, $cond_N$ 
if:
%for $D_j \subset D$ where $D_j$ is defined by the condition on $\alpha_j$, then for all data subsets $D' \subset D_j$, defined by a condition on $\alpha_i$,
\[
\frac{\# \{ k \mid \mathsf{DP}[k][n+1] = 1 \} }{\# \{\mathsf{DP}\}}\leq e^{\epsilon}\cdot \frac{\# \{ k \mid \mathsf{DN}[k][n+1] = 1\}}{\# \{\mathsf{DN}\}} + \delta.
\]
\end{definition}
\fi

\begin{definition} [Controlled Fairness]\label{def:diff_fair}
Let $D \in \mathbb{R}^{s \times n}$ be a dataset,
% with features $\{f_1, \ldots, f_n\}$.
let $cond_N$ be a set of conditions on some unprotected features in $F^N$
 and let~$cond_P$ be a condition on a protected feature in $F^P$. Let $\lnot cond_P$ denote the negation of the condition.
Let $D^+ \in \mathbb{R}^{s \times (n+1)}$ denote the \emph{labeled} dataset obtained
by running (binary) classifier $C$ on dataset $D$.

Define the datasets $\mathsf{DNP}$,  $\mathsf{\lnot DNP}$ to be the
output of the SQL ``Select'' queries  
$$\mathsf{DNP} := \sigma_{cond_N, cond_P}(D^+) \quad \mathsf{\lnot DNP} := \sigma_{cond_N, \lnot cond_P}(D^+)$$
We require that
$\#\{\mathsf{DNP}\} > 0$ and   $\#\{\mathsf{\lnot DNP}\} > 0$.

Define the ratio of a dataset as follows:
$$
\mathsf{ratio}(D) = \frac{\# \{ k \mid \mathsf{D}[k][n+1] = 1 \} }{\# \{\mathsf{D}\}}
$$

We say that C is  a fair classifier with respect to $D$, $cond_P$, $cond_N$ if:
\[
\mathsf{ratio}(\mathsf{DNP}) \approx \mathsf{ratio}(\mathsf{\lnot DNP})
\]
%for $D_j \subset D$ where $D_j$ is defined by the condition on $\alpha_j$, then for all data subsets $D' \subset D_j$, defined by a condition on $\alpha_i$,
%\[
%\frac{\# \{ k \mid \mathsf{DNP}[k][n+1] = 1 \} }{\# \{\mathsf{DNP}\}} \approx 
%\frac{\# \{ k \mid \mathsf{\lnot DNP}[k][n+1] = 1\}}{\# \{\mathsf{\lnot DNP}\}}.
%\]
\end{definition}

\iffalse
\begin{definition} [Risk Calculation]\label{def:risk}
Let $D \in \mathbb{R}^{s \times n}$ be a dataset and let $A$ be a risk assignment algorithm. Algorithm $A$ takes as input a data point $D^i \in \mathbb{R}^n$ and returns real-value score $A(D^i) \in [0,1]$. We use the notation $A(D)$ (without superscript $i$) to represent a vector of risk computed for each data point of dataset $D$. 
\end{definition}
\fi
\iffalse
\anote{just wanted to bold the following ...}
To-Do Notes: Compare this proposed definition to existing definitions, issue of correlated features(?)

Add the picture of the two distributions and the threshold line, and the idea of moving the distributions towards each other.  Try to keep the volume constant and achieve parity on one side of the threshold line. 

We note that this definition is similar to ``conditional parity'' proposed in \cite{ritov2017conditional} \placeholder{describe CP}, though our proposal focuses on evaluating the classifier on the dataset as opposed to considering parity of a random variable.  
\fi

%\section{Achieving Controlled Fairness}\label{sec:controlled_fairness}

%In this section, 
We next introduce two algorithms
for achieving controlled fairness.
We then implement and experimentally validate
our algorithms on multiple datasets.

Let $D_{(1)}$ be a dataset
and let $A_{(1)}$ be a risk assignment algorithm trained on this dataset. $A^C_{(1)}$ is the corresponding classifier, which 
we assume does not achieve the controlled fairness notion as it was defined in Definition~\ref{def:diff_fair}.

%Given a new dataset $D_{(2)}$ we first compute its labeled version $D_{(2)}^{+}$ using classifier $A_{(1)}$ to generate the labels. Then we filter the data using the unprotected features and divide the dataset into the protected features and unprotected feature.  The two sets can be computed in a similar fashion as in Definition~\ref{def:diff_fair} as follows,

%$$\mathsf{DNP_{(2)}} := \sigma_{cond_P, cond_N}(D_{(2)}^+); \quad \mathsf{\lnot DNP_{(2)}} := \sigma_{\lnot cond_P, cond_N}(D_{(2)}^+).$$

%Since the classifier $A_{(1)}$ is potentially unfair the following equation will not necessarily hold

Specifically, given a new dataset $D_{(2)}$, 
we label the dataset with the output of the classifer 
$A^C_{(1)}$ to obtain
$D^+_{(2)}$.
Since classifier $A_{(1)}^C$ does not achieve the controlled fairness notion (see Definition~\ref{def:diff_fair}) with respect to $D_{(2)}$
(if it did, we would be done) we have that

\[
\mathsf{ratio}(\mathsf{DNP_{(2)}}) \not\approx \mathsf{ratio}(\mathsf{\lnot DNP_{(2)}}),
\]

where
$$\mathsf{DNP}_{(2)} := \sigma_{cond_N, cond_P}(D^+_{(2)}) \quad \mathsf{\lnot DNP_{(2)}} := \sigma_{cond_N, \lnot cond_P}(D^+_{(2)})$$

%In the following WLOG we assume that $\mathsf{ratio}(\mathsf{DNP_{(2)}}) \geq \mathsf{ratio}(\mathsf{\lnot DNP_{(2)}})$.

Our goal is to remove the bias from $A_{(1)}^C$,
even though we do not have access to unbiased training data. In the following two subsections we present two algorithms to solve this problem.
Specifically, we would like to leverage
$A_{(1)}^C$ to train
a new classifier $C_{(2)}$ that remains accurate but
is unbiased.
The idea of the following algorithms are to use
$A_{(1)}^C$ to selectively relabel the dataset $D_{(2)}^{+}$ such that the biased is removed. 
We denote the relabeled dataset by $D^+_{(3)}$
and we refer to $D^+_{(3)}$
as a \emph{synthetic} dataset,
since the labels of $D^+_{(3)}$
do not correspond to either the original labels
or the labels produced by a classifier.
Specifically, we will consider two ways of 
constructing $D^+_{(3)}$ such that the following is satisfied:

\[
\mathsf{ratio}(\mathsf{DNP_{(3)}}) \approx \mathsf{ratio}(\mathsf{\lnot DNP_{(3)}}),
\]

where
$$\mathsf{DNP}_{(3)} := \sigma_{cond_N, cond_P}(D^+_{(3)}) \quad \mathsf{\lnot DNP_{(3)}} := \sigma_{cond_N, \lnot cond_P}(D^+_{(3)})$$

We then train a new classifier $C_{(2)}$ on the synthetic dataset.
We expect $C_{(2)}$ to achieve the controlled
fairness definition
with respect a newly sampled dataset $D_{(4)}$ which has never been seen by the classifier, while accuracy
with respect to the true labels remains high.
We will validate these expectations with our 
experimental results in Section~\ref{sec:exp}.
In the following, we describe two
algorithms for generating the synthetic labeled dataset
$D^+_{(3)}$.

\subsection{Algorithm 1: Synthetic Data via Selective Risk Adjustment} \label{sec:alg1}

In this section we present our first proposed algorithm.
The idea of this algorithm is to adjust the risk values associated with one of the protected classes, compute new
labels based on the adjusted risk values and output the resulting
database as $D^+_{(3)}$.
%to satisfy fairness as defined in Definition~\ref{def:diff_fair}.
%In this section we introduce a parameter called $\Delta$ which can be set in order to fix the bias in the biased classifier $A$.
Recall that $A_{(1)}$ is trained on dataset $D_{(1)}$ and that labeled dataset $D_{(2)}^{+}$ is obtained by applying classifier $A_{(1)}^C$ on a new dataset $D_{(2)}$.
Specifically,
the labeled dataset is computed as follows:
\[
    D_{(2)}^{+}[i][n+1] = 
\begin{cases}
    1,& \text{if } A_{(1)}(D_{(2)}^i) \geq \mathsf{threshold}\\
    0,              & \text{otherwise}
\end{cases}
\]

For the case of logistic regression the $\mathsf{threshold}$ in the above equation is usually set at $0.5$. 

Since we assumed classifier $A_{(1)}^{C}$ does not
already achieve the controlled fairness definition,
we have WLOG that
$\mathsf{ratio}(\mathsf{DNP_{(2)}}) \geq \mathsf{ratio}(\mathsf{\lnot DNP_{(2)}})
$. In particular:

\begin{equation}
\label{eqn:bias1}
    \frac{\# \{ k \mid A_{(1)}(\mathsf{DNP}_{(2)}^k) \geq \mathsf{threshold} \} }{\# \{\mathsf{DNP}_{(2)}\}} \geq \frac{\# \{ k \mid A_{(1)}(\lnot \mathsf{DNP}_{(2)}^k) \geq \mathsf{threshold} \} }{\# \{\lnot \mathsf{DNP}_{(2)}\}}  
\end{equation}

Let 
\begin{equation*}
    \alpha := \frac{\# \{\mathsf{DNP}_{(2)}\} \cdot \# \{ k \mid A_{(1)}(\lnot \mathsf{DNP}_{(2)}^k) \geq \mathsf{threshold} \} }{\# \{\lnot \mathsf{DNP}_{(2)}\}}
\end{equation*}

To obtain synthetic dataset $D^+_{(3)}$,
we first compute $\Delta$ such that the following holds:

\begin{equation}
    \# \{ k \mid \big( A_{(1)}(\mathsf{DNP}_{(2)}^k) - \Delta \big) \geq \mathsf{threshold} \} \approx \alpha 
\label{eqn:set_delta}
\end{equation}

\begin{figure}[ht]
  \centering
  \includegraphics[width=0.5\textwidth]{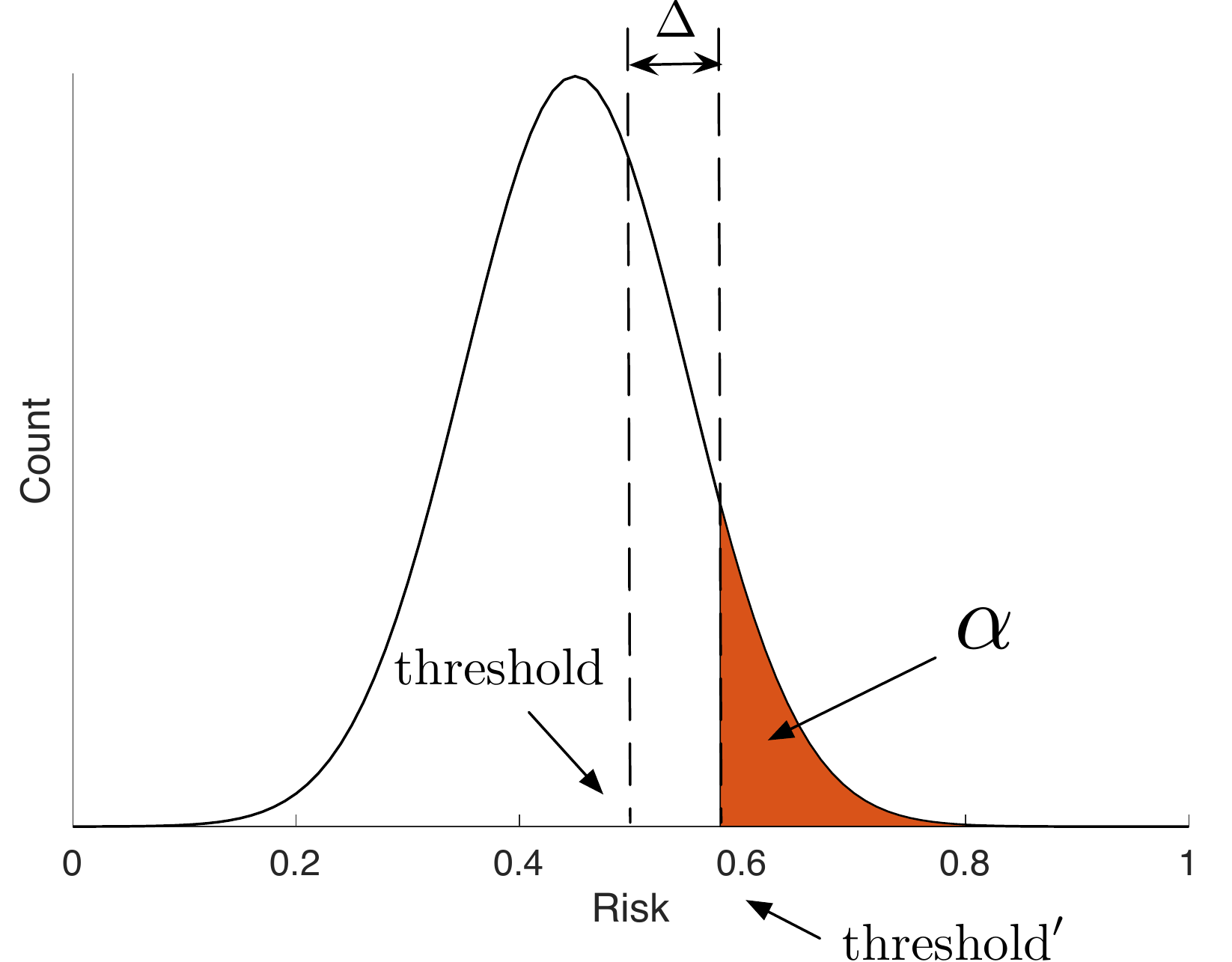}
  \caption{
  Note that the quantity $\# \{ k \mid  A_{(1)}(\mathsf{DNP}_{(2)}^k)  \geq \mathsf{threshold}' \}$ is represented by the area under the curve and to the right of the vertical line passing through $(\mathsf{threshold}', 0)$.
  Therefore, we set the value of $\mathsf{threshold}'$ so that the area of the marked region is equal to $\alpha$. Then $\Delta$ is
  set to $\mathsf{threshold}' - \mathsf{threshold}$.
}
  \label{fig:set_delta}
\end{figure}

In order to compute $\Delta$ we sort 
$\mathsf{DNP}_{(2)}$ 
according to the risk value outputted by $A_{(1)}$
on each entry
and find the maximal value of $\mathsf{threshold'}$ such that $\# \{ k \mid  A_{(1)}(\mathsf{DNP}_{(2)}^k)  \geq \mathsf{threshold'} \} \approx \alpha$. Then $\Delta$ can be computed as $\Delta = \mathsf{threshold}' - \mathsf{threshold}$. Figure~\ref{fig:set_delta} shows an example of the distribution of $A_{(1)}(\mathsf{DNP}_{(2)})$ and pictorially represents the method of finding $\Delta$. 

%\begin{figure}[h!]
%  \centering
%  \includegraphics[width = 8.5 cm]{Figure/NYPD/apply_delta.pdf}
%  \caption{Visualization of applying the optimal delta to equalize white and nonwhite probabilities}
%\end{figure}

$\mathsf{DNP}_{(3)}$ is equivalent to
$\mathsf{DNP}_{(2)}$, except for the final
column (the $(n+1)$-st column), which corresponds to the labels.
The labels of $\mathsf{DNP}_{(3)}$ are defined as follows:

\[
    \mathsf{DNP}_{(3)}[i][n+1] = 
\begin{cases}
    1,& \text{if } A_{(1)}(\mathsf{DNP}_{(2)}^i) - \Delta  \geq \mathsf{threshold}\\
    0,              & \text{otherwise}
\end{cases}
\]

It is straightforward to see that the following property is satisfied

\begin{equation*}\label{eqn:risk_flip1}
  \frac{ \#\{ k \mid \mathsf{DNP}_{(3)}[k][n+1] = 1 \}}{\# \{\mathsf{DNP}_{(3)}\}} \approx \frac{\# \{ k \mid \lnot \mathsf{DNP}_{(2)}[k][n+1] = 1 \} }{\# \{ \lnot \mathsf{DNP}_{(2)}\}}.  
\end{equation*}

$D_{(3)}^+$ is then defined to be a concatenation of the following datasets. 

\begin{equation*}
D_{(3)}^{+} := \mathsf{DNP}_{(3)} \quad | \quad \lnot \mathsf{DNP}_{(2)} \quad | \quad \sigma_{\lnot cond_N}(D_{(2)}^+)    
\end{equation*}

The synthetic dataset $D_{(3)}^{+}$ will be used to train a new classifier $C_{(2)}$. Algorithm~\ref{alg:alg1} shows pseudocode for this algorithm. The \textsc{Filter} function takes a dataset $D$, and returns the rows which satisfies whatever conditions $cond$ is passed to it as input. The \textsc{CheckFair} function takes two dataset and the label column and check whether the fairness notion, as it is introduced in Definition~\ref{def:diff_fair}, is satisfied. The \textsc{CheckFair} function returns $1$ if fairness notion is achieved and returns $0$, otherwise. 

\begin{algorithm} 
%\begin{breakablealgorithm}
\caption{Synthetic Data via Selective Risk Adjustment} \label{alg:alg1}
\begin{algorithmic}[1]
\Procedure {main}{$D,A,colnum,rownum,th, cond_P, cond_N$}
\State Call \textsc{RiskAssignment}($D, A, colnum, rownum$)
\State Call \textsc{InitializeLabels}($D, colnum, rownum, th$)
\State Call \textsc{SyntheticGen}($D, colnum, rownum,th, cond_P, cond_N$)
\State $C \leftarrow$ \textsc{Learn}($D, colnum+1, rownum$)
\State Output $C$
\EndProcedure

\vspace{2mm}

\Procedure {RiskAssignment}{$D, A, colnum,rownum$}
	\For {$i = 1$ to $rownum$}
		\State {Set $D[i][colnum+2] = A(D[i][1], \ldots, D[i][colnum])$}
	\EndFor
\EndProcedure

\vspace{2mm}

\Procedure {InitializeLabels}{$D,colnum,rownum,th$}
	\For {$i = 1$ to $rownum$}
	    \If{$D[i][colnum+2] \geq th$}
		\State {Set $D[i][colnum+1] =1$}
		\Else 
		\State {Set $D[i][colnum+1] = 0$}
		\EndIf
	\EndFor
\EndProcedure

\vspace{2mm}

\Procedure {SyntheticGen}{$D,colnum,rownum,th, cond_P, cond_N$}
	\State Set $\mathsf{DNP} = $\textsc{Filter}$(D,cond_N, cond_P)$
	\State Set $\lnot \mathsf{DNP} = $\textsc{Filter}$(D, cond_N,\lnot cond_P)$
	\State Set $\lnot \mathsf{DN} = $\textsc{Filter}$(D, \lnot cond_N)$
	\If{\textsc{CheckFair}$(\mathsf{DNP}, \lnot \mathsf{DNP}, colnum+1) = 0$}
	\State sum = 0
	\For {$i = 1$ to $\#\{\lnot \mathsf{DNP}\}$}
		\If{$\lnot \mathsf{DNP}[i][colnum+2] \geq th$}
		    \State sum = sum + 1
		\EndIf
	\EndFor
	\State $\alpha = \frac{\#\{\mathsf{DNP}\} \cdot \text{sum}}{\#\{\lnot \mathsf{DNP}\}}$
	\State Sort $\mathsf{DNP}$ according to $colnum+2$ from largest to smallest
	\State $th' = \mathsf{DNP}[\text{round}(\alpha)][colnum+2]$
	\State $\Delta = th' - th$
	\For {$i = 1$ to $\#\{\mathsf{DNP}\}$}
		\State $\mathsf{DNP}[i][colnum+2] = \mathsf{DNP}[i][colnum+2] - \Delta$
		%\If{$\mathsf{DNP}[i][colnum+2] \geq th$}
		%    \State $\mathsf{DNP}[i][colnum+1] = 1$
		%\Else
		%    \State $\mathsf{DNP}[i][colnum+1] = 0$
		%\EndIf
	\EndFor
	\State Call \textsc{InitializeLabels}($\mathsf{DNP}, colnum, \#\{\mathsf{DNP}\}, th$)
    \State $D := \text{Concatenate}(\mathsf{DNP},  \lnot \mathsf{DNP}, \lnot \mathsf{DN})$ 
	\EndIf
	
\EndProcedure
\end{algorithmic}
\end{algorithm}

\subsection{Algorithm 2: Synthetic Data via Risk Based Flipping} \label{sec:alg2}

In this section, we present our second proposed
algorithm.
The idea of this algorithm is to preserve the original labels of the datapoints in $D_{(2)}$
(as opposed to using $A_{(1)}^C$ to fully relabel the dataset)
and flip only the minimal number
of datapoints within one of the protected classes to construct $D^+_{(3)}$
that satisfies
\[
\mathsf{ratio}(\mathsf{DNP_{(3)}}) \approx \mathsf{ratio}(\mathsf{\lnot DNP_{(3)}}).
\]
$A_{(1)}$ will be 
utilized to provide a risk score that helps decide which datapoints
should be flipped.

Recall that as in the previous sections classifier $A_{(1)}^C$ is trained on dataset $D_{(1)}$. In this section, we assume a new \emph{labeled} dataset $D^+_{(2)}$ is given, labeled
with the true labels.
If labeled dataset $D^+_{(2)}$ is not biased,
then we can trivially set $D^+_{(3)} := D^+_{(2)}$.
Therefore,
we assume WLOG that the following holds

$$
\frac{\# \{ k \mid \mathsf{DNP}_{(2)}[k][n+1] = 1 \} }{\# \{\mathsf{DNP}_{(2)}\}} \geq \frac{\# \{ k \mid \lnot \mathsf{DNP}_{(2)}[k][n+1] = 1 \} }{\# \{ \lnot \mathsf{DNP}_{(2)}\}}.
$$

Similar to the previous algorithm we fix the right hand side:

% \# \{ k \mid \mathsf{DNP}_{(3)}^{+}[k][n+1] = 1 \}
\begin{equation*}\label{eqn:risk_flip2}
 \alpha := \frac{\# \{\mathsf{DNP}_{(2)}\} \cdot \# \{ k \mid \lnot \mathsf{DNP}_{(2)}[k][n+1] = 1 \} }{\# \{ \lnot \mathsf{DNP}_{(2)}\}}    
\end{equation*}

The above indicates that we should construct $DNP_{(3)}$ 
such that it has $\alpha$ number of rows with label 1. Since there are currently $\# \{ k \mid \mathsf{DNP}_{(2)}[k][n+1] = 1 \}$ rows in $\mathsf{DNP_{(2)}}$ with label 1, we must flip the labels for $\beta = \alpha - \# \{ k \mid \mathsf{DNP}_{(2)}[k][n+1] = 1 \}$ number of rows from 1 to 0. To do so, we first divide $\mathsf{DNP_{(2)}}$ into disjoint datasets based on its labels and flip enough samples only from the dataset with label of 1. Let $cond_L$ be the condition that selects the datapoints 
of $\mathsf{DNP_{(2)}}$ 
with label of 1 and $\lnot cond_L$ be the condition that selects datapoints with label of 0. Then we construct the following two datasets from $\mathsf{DNP_{(2)}}$,
$$
\mathsf{DNPL}_{(2)} := \sigma_{cond_N, cond_P, cond_L}(D_{(2)}^{+}) \quad \mathsf{\lnot DNPL}_{(2)} := \sigma_{cond_N, cond_P, \lnot cond_L}(D_{(2)}^{+})
$$
Then, we sort $\mathsf{DNPL}_{(2)}$ according to the score of each data point, as assigned by $A_{(1)}$, from smallest to largest. 
$\mathsf{DNPL}_{(3)}$ is equivalent to
(the sorted version of)
$\mathsf{DNPL}_{(2)}$, except for the final
column (the $(n+1)$-st column), which corresponds to the labels.
The labels of $\mathsf{DNPL}_{(3)}$ are defined as follows:
\[
    \mathsf{DNPL}_{(3)}[i][n+1] = 
\begin{cases}
    0,& \text{if } i \leq \beta\\
    1,              & \text{otherwise}
\end{cases}
\]

We then define $\mathsf{DNP}_{(3)}$ as follows: $\mathsf{DNP}_{(3)} := \mathsf{DNPL}_{(3)} | \lnot \mathsf{DNPL}_{(2)}$ and it is straightforward to see that the following property is satisfied

\begin{equation*}\label{eqn:risk_flip3}
  \frac{ \#\{ k \mid \mathsf{DNP}_{(3)}[k][n+1] = 1 \}}{\# \{\mathsf{DNP}_{(3)}\}} \approx \frac{\# \{ k \mid \lnot \mathsf{DNP}_{(2)}[k][n+1] = 1 \} }{\# \{ \lnot \mathsf{DNP}_{(2)}\}}.  
\end{equation*}

Similar to the previous algorithm, we define the synthetic dataset $D_{(3)}^+$ to be a concatenation of the following datasets. 

$$
D_{(3)}^{+} := \mathsf{DNP}_{(3)} \quad | \quad \lnot \mathsf{DNP}_{(2)} \quad | \quad \sigma_{\lnot cond_N}(D_{(2)}^+)
$$

The new dataset $D_{(3)}^{+}$ will be used to train a new classifier $C_{(2)}$. Algorithm~\ref{alg:alg2} shows pseudocode for this algorithm. The \textsc{Filter} function takes a dataset $D$, and returns the rows which satisfies whatever conditions $cond$ is passed to it as input. The \textsc{CheckFair} function takes two dataset and the label column and check whether the fairness notion, as it is introduced in Definition~\ref{def:diff_fair}, is satisfied. The \textsc{CheckFair} function returns $1$ if fairness notion is achieved and returns $0$, otherwise.

\begin{algorithm}
\caption{Synthetic Data via Risk Based Flipping} \label{alg:alg2}
\begin{algorithmic}[1]
\Procedure {main}{$D,A,colnum,rownum,th, cond_P, cond_N, cond_L$}
\State Call \textsc{RiskAssignment}($D, A, colnum, rownum$)
\State Call \textsc{SyntheticGen}($D, colnum, rownum,th, cond_P, cond_N, cond_L$)
\State $C \leftarrow$ \textsc{Learn}($D, colnum+1, rownum$)
\State Output $C$
\EndProcedure

\vspace{2mm}

\Procedure {RiskAssignment}{$D, A, colnum,rownum$}
	\For {$i = 1$ to $rownum$}
		\State {Set $D[i][colnum+2] = A(D[i][1], \ldots, D[i][colnum])$}
	\EndFor
\EndProcedure

%\vspace{2mm}

%\Procedure {Relabels}{$D,colnum,rownum,th$}
%	\For {$i = 1$ to $rownum$}
%	    \If{$D[i][colnum+2] \geq th$}
%		\State {Set $D[i][colnum+1] =1$}
%		\Else 
%		\State {Set $D[i][colnum+1] = 0$}
%		\EndIf
%	\EndFor
%\EndProcedure

\vspace{2mm}

\Procedure {SyntheticGen}{$D,colnum,rownum,th, cond_P, cond_N, cond_L$}
	\State Set $\mathsf{DNP} = $\textsc{Filter}$(D,cond_N, cond_P)$
	\State Set $\mathsf{DNPL} = $\textsc{Filter}$(D, cond_N, cond_P, cond_L)$
	\State Set $\lnot \mathsf{DNPL} = $\textsc{Filter}$(D, cond_N, cond_P, \lnot cond_L)$
	\State Set $\lnot \mathsf{DNP} = $\textsc{Filter}$(D, cond_N,\lnot cond_P)$
	\State Set $\lnot \mathsf{DN} = $\textsc{Filter}$(D, \lnot cond_N)$
	\If{\textsc{CheckFair}$(\mathsf{DNP}, \lnot \mathsf{DNP}, colnum+1) = 0$}
	\State sum = 0
	\For {$i = 1$ to $\#\{\lnot \mathsf{DNP}\}$}
		\If{$\lnot \mathsf{DNP}[i][colnum+1] == 1$}
		    \State sum = sum + 1
		\EndIf
	\EndFor
	\State $\alpha = \frac{\#\{\mathsf{DNP}\} \cdot \text{sum}}{\#\{\lnot \mathsf{DNP}\}}$
	\State $\beta = \alpha - \#\{ \mathsf{DNPL}\}$
	\State Sort $\mathsf{DNPL}$ according to $colnum+2$ from smallest to largest
	\For {$i = 1$ to $\beta$}
		\State $\mathsf{DNPL}[i][colnum+1] = 0$
	\EndFor
    \State $D := \text{Concatenate}(\mathsf{DNPL}, \lnot \mathsf{DNPL},  \lnot \mathsf{DNP}, \lnot \mathsf{DN})$ 
	\EndIf
	
\EndProcedure
\end{algorithmic}
\end{algorithm}

\subsection{Choice of dataset} \label{sec:dataset}

\paragraph{Stop, Question and Frisk Dataset.}
The \emph{Stop, Question and Frisk} dataset is a publicly available dataset that consists of information collected by New York Police Department officers since 2003~\cite{sqfdata}. We selected the data from year 2012 as it had a sufficient number of entries for training purposes. The original dataset from 2012 had $532,911$ rows and $112$ columns and after cleaning $477,840$ rows remained.
Additionally, we selected $33$ relevant columns, based on the description which is provided with the dataset. 
%We note that some of the columns in the original data would not make sense to include in the model since they would be known after the output of the model was determined. For example, if we focus on predicting whether an individual will be frisked, a column which represents whether they carry a hand gun should not be taken into account. 
Each row of the dataset represents an individual who has been stopped by an officer and includes detailed information about the incident such as time of stop, reason for stop, crime they are suspected of, etc. 

\begin{figure}[ht]
\begin{minipage}[t]{.45\textwidth}
    \centering
  \includegraphics[width=\linewidth]{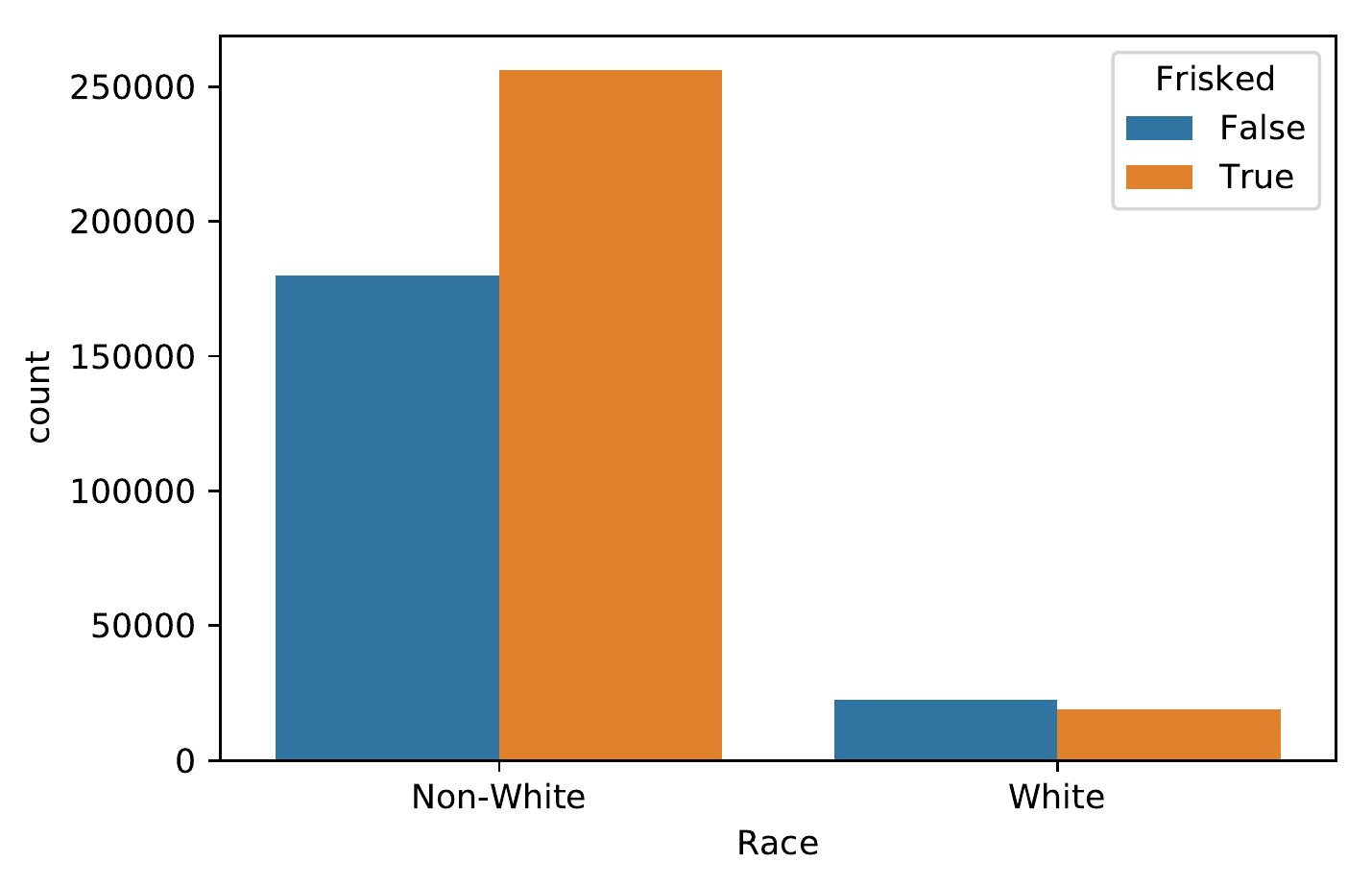}
  \caption{Number of people who got frisked in each race}
  \label{fig:sqf_race_orig}
\end{minipage}%
\hfill%
\begin{minipage}[t]{.45\textwidth}
  \centering
  \includegraphics[width=\linewidth]{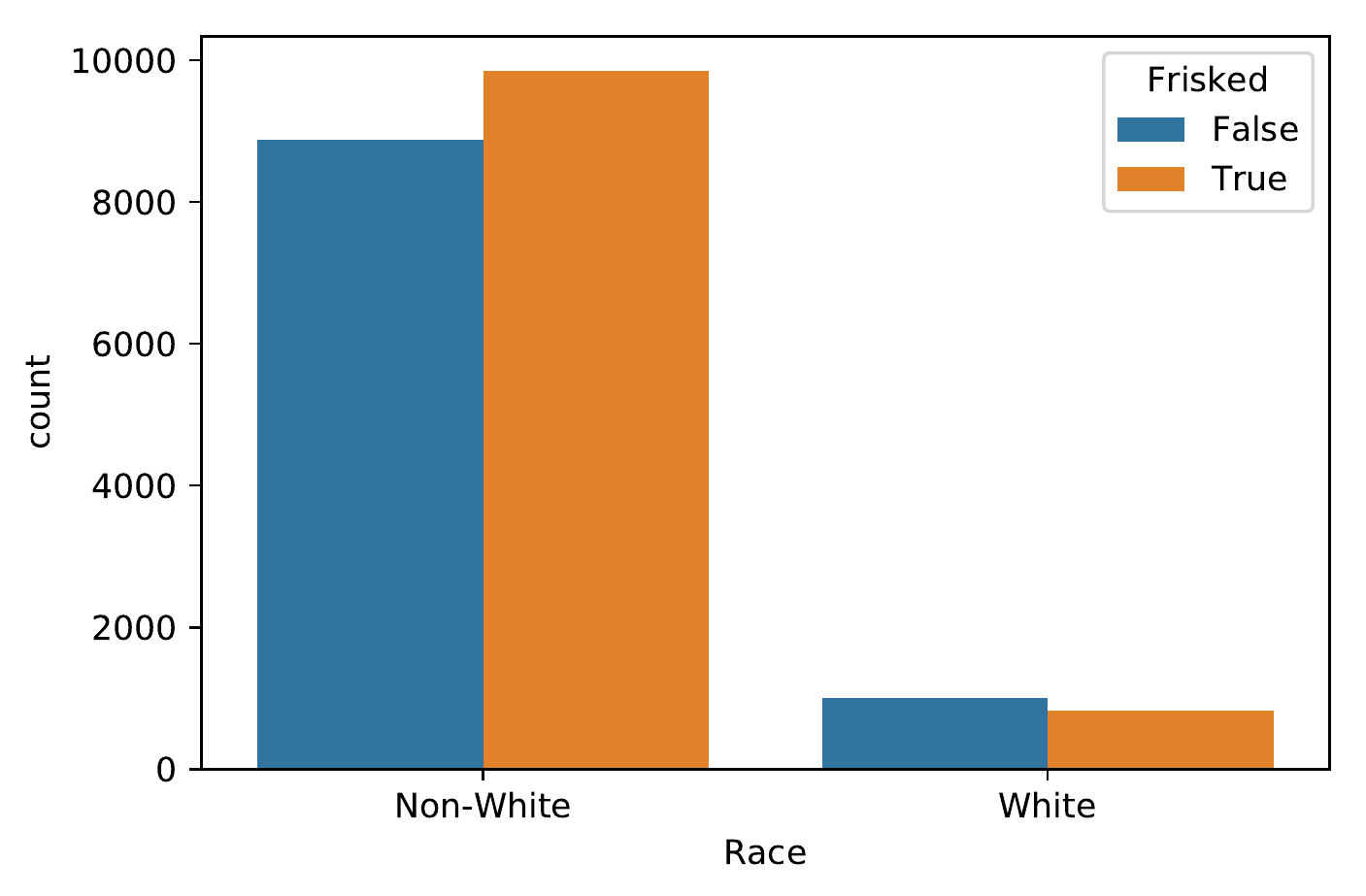}
  \caption{Number of people who got frisked conditioned on being suspected of committing an assault}
  \label{fig:sqf_race_filtered}
\end{minipage}
\end{figure}

\begin{figure}[ht]
  \centering
  \includegraphics[width =0.5\linewidth]{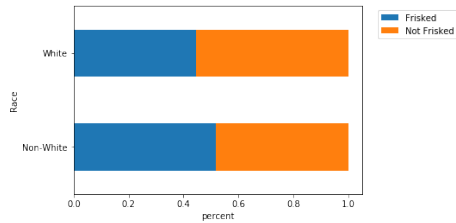}
  \caption{Comparison of frisked ratio between races conditioned on being suspected of committing an assault}
  \label{fig:sqf_filter_stackedbar}

\end{figure}

Figure~\ref{fig:sqf_race_orig} shows the majority of the people in the dataset (i.e.~a majority of people who were stopped by an officer) are non-white. In addition, a significantly higher proportion of non-whites are frisked compared to whites. 
%On the other hand, Figure~\ref{fig:sqf_outcome_orig} shows that non-whites also are also more likely to be suspected of an assault. Of course, this suspicion may also be subject to bias. 
Figure~\ref{fig:sqf_race_filtered} shows that when we filter on the type of crime an individual is suspected of, e.g.~assault vs.~non-assualt, Non-Whites are still more likely to be frisked. 
Finally, Figure~\ref{fig:sqf_filter_stackedbar} presents the percentage of frisked individuals within each race Figure~\ref{fig:sqf_race_filtered}. It can be seen that White individuals are less likely to be frisked, even when suspected of the same type of crime (assault).

%\begin{figure}[h!]
%  \centering
%  \includegraphics[width = 8.5 cm]{Figure/NYPD/race.pdf}
%  \caption{Comparison of original classification between white and nonwhite classes}
%\end{figure}

%\begin{figure}[h!]
%  \centering
%  \includegraphics[width = 8.5 cm]{Figure/NYPD/filter.pdf}
%  \caption{Comparison of filtering condition between white and nonwhite classes}
%\end{figure}

\paragraph{Adult Income Dataset.}
The \emph{Adult Income} dataset was taken from the 1994 Census Database and contains $15$ columns of demographic information collected by the census. The dataset contains $48,842$ entries. The task for this dataset was to predict whether a given individual earned more than \$50K per year. 
We chose this dataset since
such a classifier can then be used, for example, to decide
whether an individual gets approved for a loan.
Three columns were dropped from the dataset when training the models. Native country was dropped because it did not contain a lot of information, as the vast majority of individuals were from the United States, and using the column as a training feature would have added a high degree of dimensionality. Education level was also dropped because the same information was encoded in a separate numeric column. Finally, fnlwgt - the statistical weight assigned to each individual by the census - was dropped because it had extremely low correlation with the target column.

\begin{figure}[ht]
\begin{minipage}[t]{.45\textwidth}
  \includegraphics[width=\linewidth]{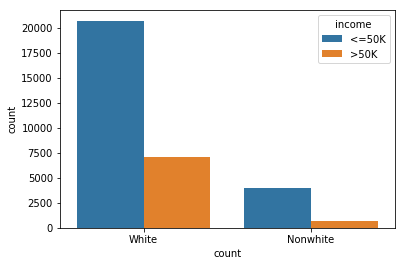}
  \caption{Number of people by race and income level}
  \label{fig:adult_race_orig}
\end{minipage}%
\hfill
\begin{minipage}[t]{.45\textwidth}
  \includegraphics[width=\linewidth]{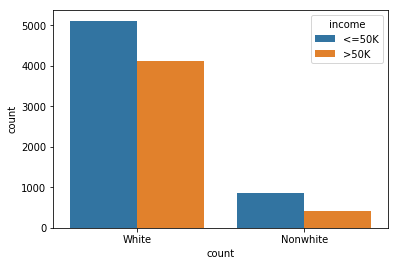}
  \caption{Number of people by race and income level conditioned on having more than 10 years of education}
  \label{fig:adult_outcome_orig}
\end{minipage}
\end{figure}

\begin{figure}[ht]
\begin{minipage}{1\textwidth}
  \centering
  \includegraphics[width=.5\textwidth]{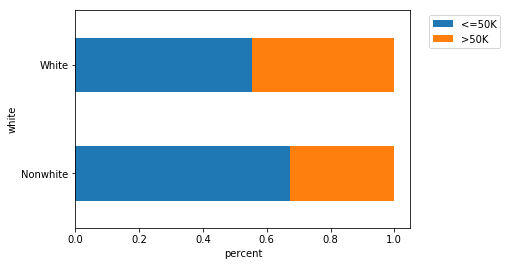}
  \caption{Comparison of ratio of income levels conditioned on having more than 10 years of education}
  \label{fig:adult_stacked_orig}
\end{minipage}
\end{figure}

Figure~\ref{fig:adult_race_orig} shows that the majority of individuals in the dataset are white, and a greater proportion of White people make more than \$50K per year than Non-whites. Figure~\ref{fig:adult_outcome_orig} shows the same plot filtered on education level (only for the subset of the dataset with more than 10 years of education). Figure~\ref{fig:adult_stacked_orig} represent the percentage for each race for the Figure~\ref{fig:adult_outcome_orig}. It can be seen that a higher percentage of White individuals versus Non-White make more than \$50K, even when controlling for education level.

\subsection{Type of models for our Algorithms}\label{sec:cl_choices}
Algorithm 1 and 2 both require two trained models,
denoted $A_{(1)}$ and $C_{(2)}$.
In this section, we will consider the choice of the
\emph{type} of model used for $A_{(1)}$ and $C_{(2)}$.
Recall that $A_{(1)}$ is a risk assignment algorithm. 
This means we have to select a type of model which outputs
continuous values representing the probability of each output class. Luckily, two of the most popular types of models ``Logistic Regression (LR)'' and ``Multi-layer Perceptron (MLP),'' have this property. While $C_{(2)}$ is only required to be a classifer, we still consider the same two types of models for it. Therefore, there are $4$ different choices for the types for $A_{(1)}/C_{(2)}$: LR/LR, LR/MLP, MLP/LR, MLP/MLP. 
As our experimental data will show, some combinations of types
perform significantly better than others, both in the case of Algorithm 1 and Algorithm 2.
In the following we will provide a high-level explanation for why both Algorithm 1 and Algorithm 2 perform better when $C_{(2)}$ is of type MLP.

For illustrative purposes, we use the \emph{Adult Income Dataset} to construct a simplified model in which only two features - age and hours per week - are considered. In this case, age is the protected feature and hours per week is the
unprotected feature.
We will observe that the synthetic dataset constructed
by Algorithm 1 and 2 is difficult for a LR-type model to learn.

\begin{figure}[ht]
\centering
\begin{minipage}{.45\textwidth}
    \centering
  \includegraphics[width=\linewidth]{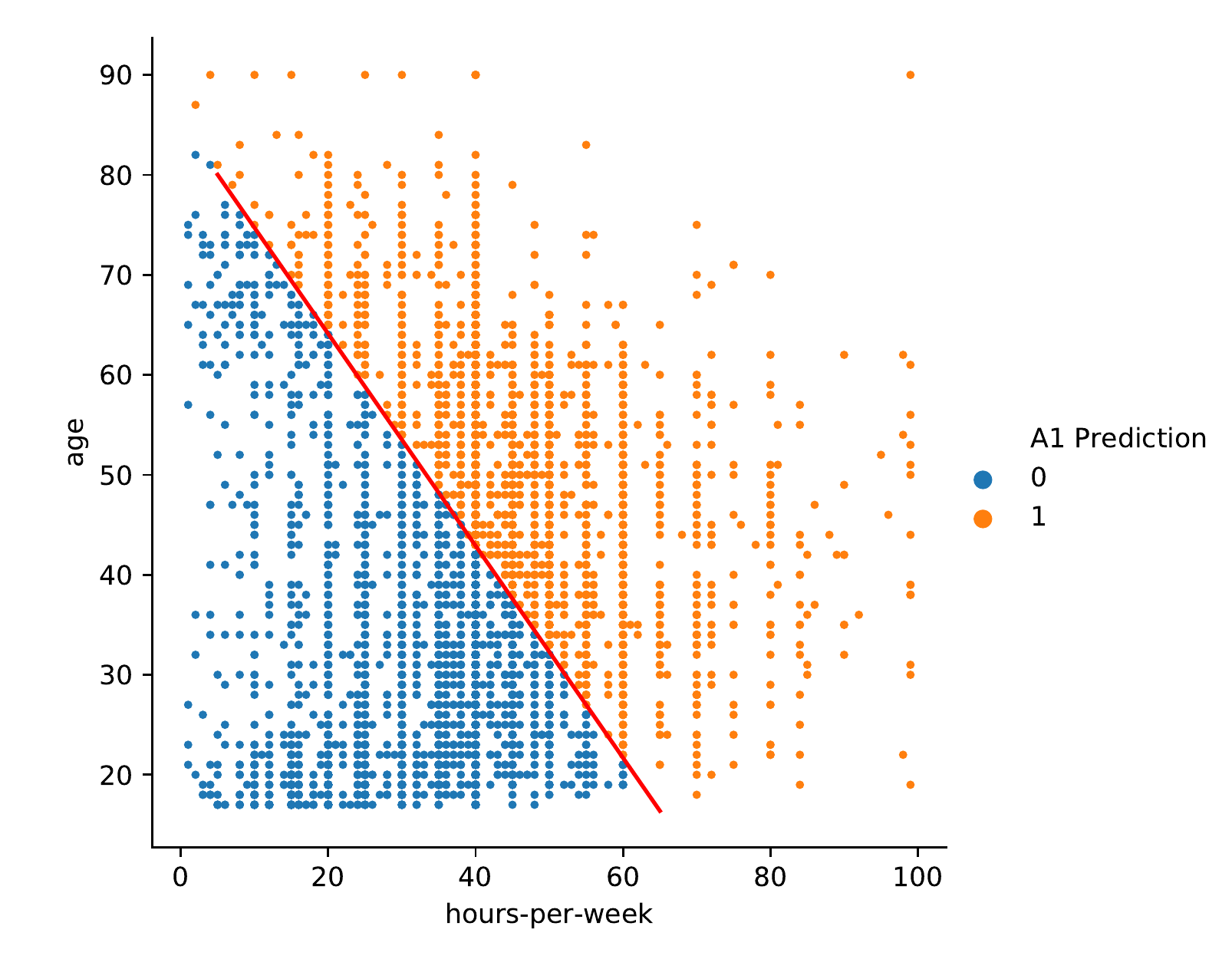}
  \caption{$A_{(1)}^C$ predictions on dataset $D_{(2)}$}
  \label{fig:adult_boundary_c1}
\end{minipage}%
\hfill
\begin{minipage}{.45\textwidth}
  \centering
  \includegraphics[width=\linewidth]{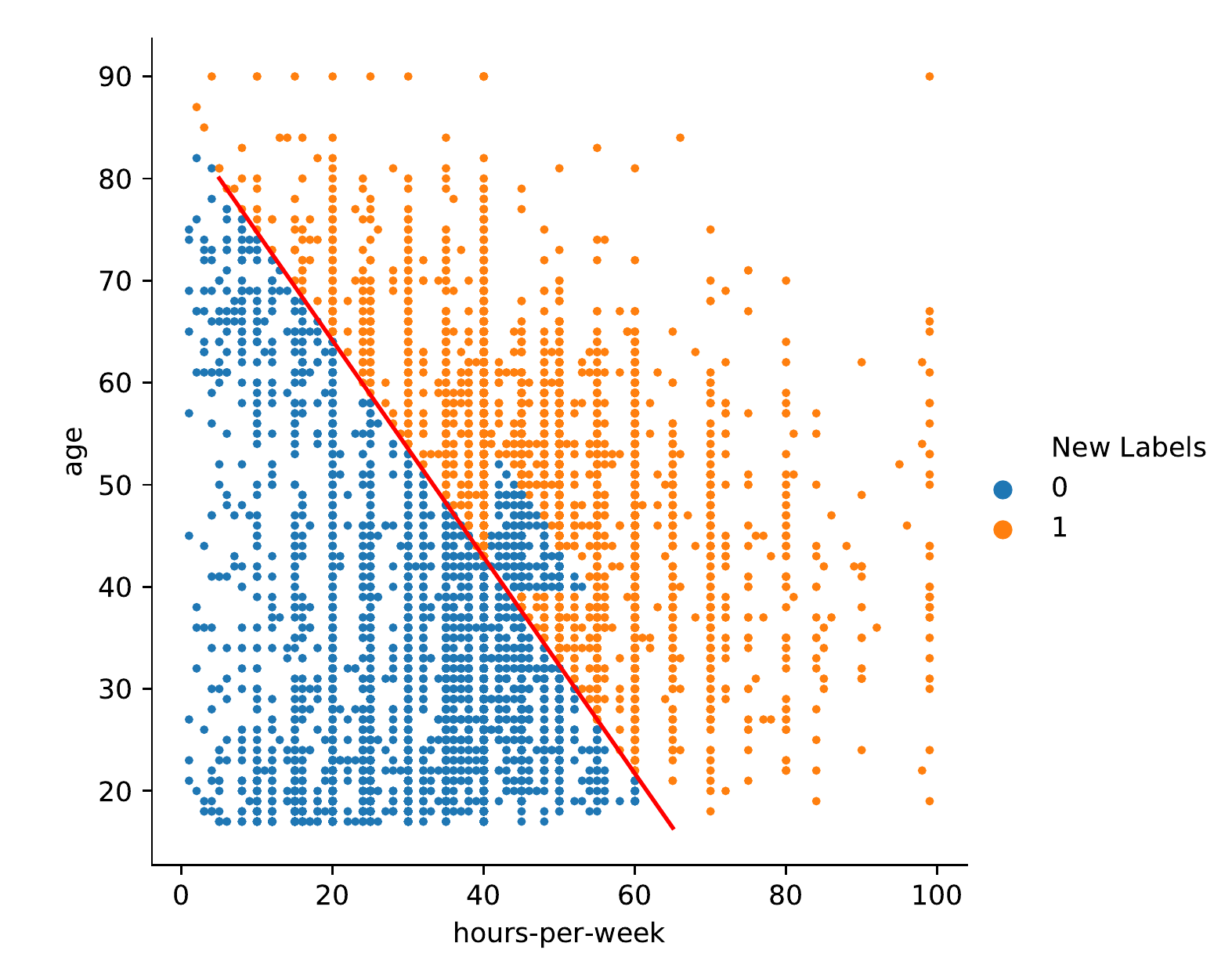}
  \caption{$A_{(1)}^C$ Predictions adjusted by applying $\Delta$ to get synthetic dataset $D_{(3)}^+$}
  \label{fig:adult_boundary_adjusted}
\end{minipage}
\end{figure}

\paragraph{\bf Algorithm 1} Recall that in this case we find an optimal value of $\Delta$ and use it to perturb the dataset
$D_{(2)}^+$, which is labeled with the output of $A_{(1)}^C$. 
Figure~\ref{fig:adult_boundary_c1} shows the prediction of the first classifier, i.e. $A_{(1)}^C$ on dataset $D_{(2)}$. The decision boundary for that classifier is also represented in the same figure. Figure~\ref{fig:adult_boundary_adjusted} shows the new labels produced after applying $\Delta$ to subset of dataset $D_{(2)}$ to generate a synthetic dataset $D_{(3)}^{+}$. Note that applying $\Delta$ creates non linear decision boundaries
(observe the blue triangular region that appears), resulting in a dataset that cannot be learned to high accuracy with logistic regression.

\begin{figure}[ht]
\centering
\begin{minipage}{0.45\textwidth}
    \centering
  \includegraphics[width=\linewidth]{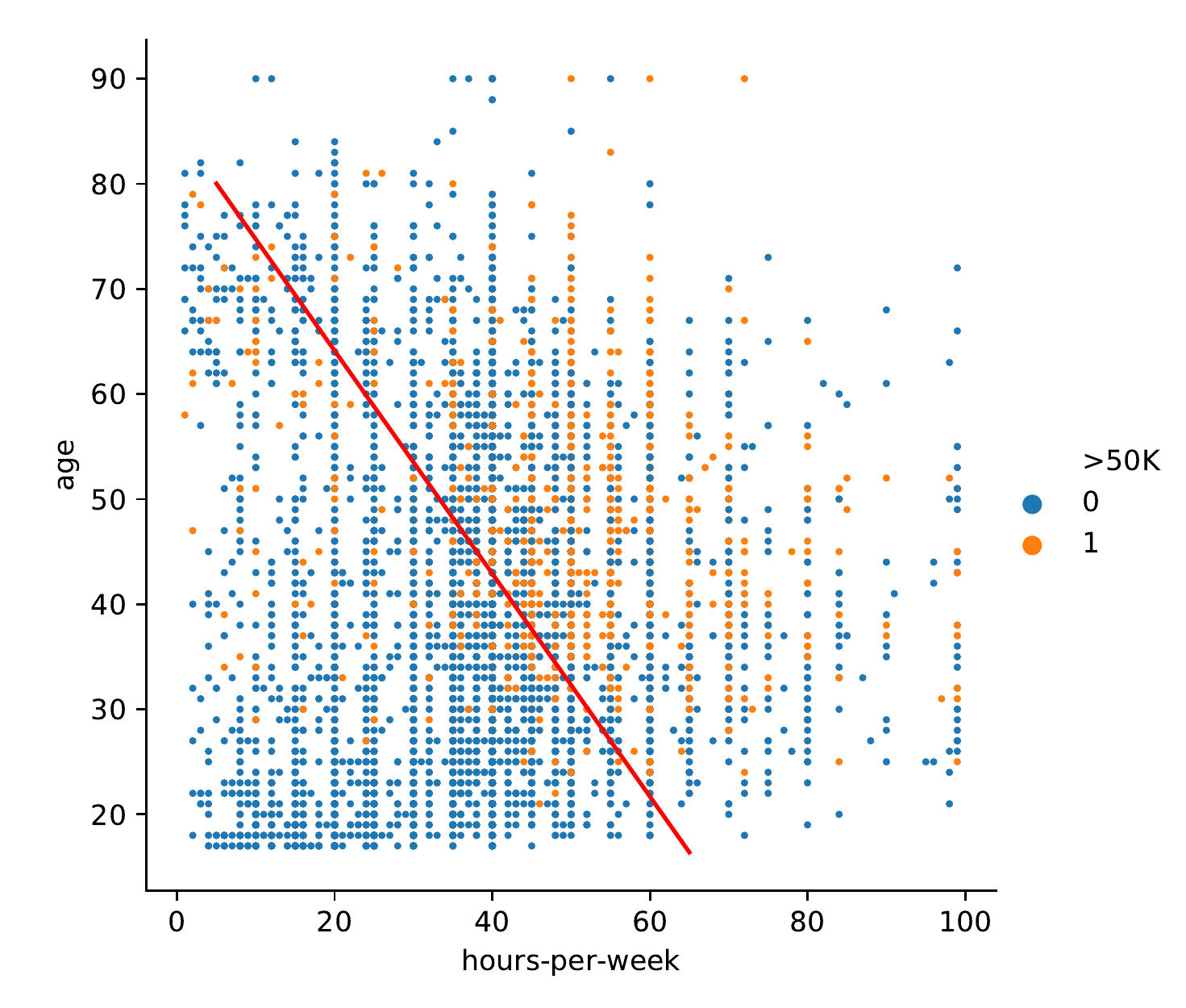}
  \caption{$A_{(1)}^C$ decision boundary with original labels of $D_{(2)}$}
  \label{fig:adult_boundary_c1_orig}
\end{minipage}%
\hfill
\begin{minipage}{.45\textwidth}
  \centering
  \includegraphics[width=\linewidth]{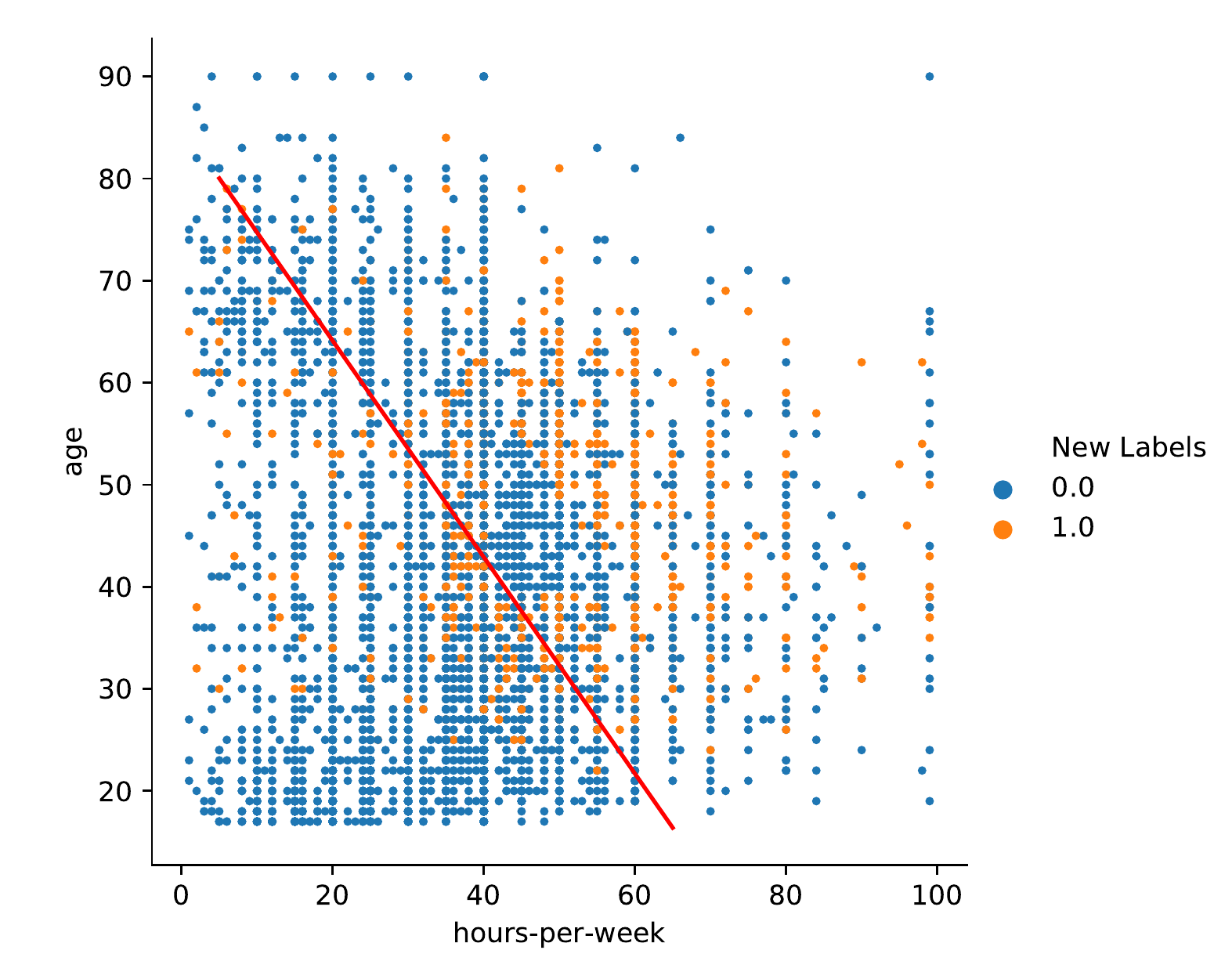}
  \caption{$A_{(1)}^C$ decision boundary and flipping labels of $D_{(2)}$ to get synthetic dataset $D_{(3)}^+$}
  \label{fig:adult_label_flipping_c1}
\end{minipage}
\end{figure}

\paragraph{\bf Algorithm 2} Recall that in this case we keep the original labels of dataset $D^+_{(2)}$ and flip only a minimal number of labels (using $A_{(1)}$ to decide which labels to flip). Figure~\ref{fig:adult_boundary_c1_orig} shows the decision boundary of $A_{(1)}^C$ as well as original labels of $D^+_{(2)}$. Figure~\ref{fig:adult_label_flipping_c1} shows the new labels generated by flipping enough sample points in the subset of dataset $D^+_{(2)}$ to generate synthetic dataset $D_{(3)}^{+}$. Similar to the previous algorithm, flipping these labels creates non linear decision boundaries (a blue triangular region again appears, although fainter than before), resulting in a dataset that cannot be learned to high accuracy with logistic regression.\\

It is therefore crucial for the second classifier to be able to find these non-linear boundaries as well. As supported by the experimental data, Multi-layer Perceptron, which potentially has non-linear decision boundaries, is therefore a good choice
for the type of the second classifer $C_{(2)}$.

\subsection{Experimental Result}\label{sec:exp}

In this section we present the result of both of the algorithms explained earlier. In each of the tables there are protected features shown in the first column. For both of the datasets we chose to use ``race'' as a protected feature, with ``W'' representing White and ``NW'' representing non-White. The unprotected features and corresponding conditions, shown by ``Filter'' in the Table, are different for each dataset. For the \emph{Stop, Question and Frisk} dataset,
the unprotected feature is \emph{type of crime}
and the corresponding condition is whether an individual is suspected of committing an assault, represented by ``A'' or not suspected of assault, represented by ``NA.''
For the \emph{Adult Income} dataset, the unprotected feature is
\emph{education level} and the corresponding condition is whether an individual has more than $10$ years of education, represented by ``E'' or not,
represented by ``UE'' in the tables. We also report the accuracy of each trained model, represented by ``Acc'' in the tables, in terms of the Area under Curve (AUC) of each model. The accuracy is measured with respect to the original labels. In each case we report the $\mathsf{ratio}$, as it is introduced in Definition~\ref{def:diff_fair}, of each dataset given the conditions on its columns. The final goal is to have a classifier which achieves fairness notion as it was introduced in Definition~\ref{def:diff_fair}.
Specifically, $\mathsf{ratio}$ should
be approximately the same
%once we filter on unprotected features and look the $\mathsf{ratio}$ 
across the protected classes. For the choice of model type, as explained in Section~\ref{sec:cl_choices}, we consider $4$ different combinations. 

To measure the performance of our algorithms, we divide our dataset into $3$ batches of size $40\%, 40\%, 20\%$. The first $40\%$ of dataset is $D_{(1)}$ and is used to construct the first classifier $A_{(1)}^C$. The second $40\%$ of dataset is $D_{(2)}$ and is used to first check the fairness notion for $A_{(1)}^C$ and construct the new dataset, i.e. $D_{(3)}^{+}$. The last $20\%$ is $D_{(4)}$ which is being used for testing the fairness of both $A_{(1)}^C$ and $C_{(2)}$. All the numbers we report correspond to the performance of the classifier with respect to the dataset $D_{(4)}$, which has never been seen by the classifier.  

\paragraph{\bf Stop, Question and Frisk Dataset} Table~\ref{tbl:nyc_lr} shows the performance---fairness and accuracy---for Algorithm 1 and Algorithm 2 on the \emph{Stop, Question and Frisk Dataset} when the first model, $A_{(1)}$, is of type \emph{Logistic Regression}. The first $2$ columns are obtained from the original labels of the data. They show 
that when we filter on ``Assault'', the $\mathsf{ratio}$ of individuals who got frisked
is $45\%$ and $52\%$ for White
and Non-White, respectively---a gap of
$7\%$.
The next $2$ columns are obtained
from the labels generated by
$A^C_{(1)}$ on the dataset.
In this case, when we filter on ``Assault'', the $\mathsf{ratio}$ of frisked individuals
is $25\%$ and $38\%$ for White and Non-White, respectively---a gap of $13\%$. By selecting the second model, $C_{(2)}$, to be of type \emph{Logistic Regression},
the ratios become
$20\%$ and $27\%$ for Algorithm 1
and $19\%$ and $25\%$ for Algorithm $2$.
Note that while
the difference in the ratios is not significantly reduced (as expected when using a second model of type LR). Once we choose MLP as the type of the second model, $C_{(2)}$, the 
ratios become $25\%$ and $27\%$ for Algorithm 1 and
$42\%$ and $46\%$ for Algorithm $2$,
reducing the gap to $2\%$ and $4\%$, respectively.
Note that the gap is comparable
for the two algorithms, Algorithm $2$ is preferable, since the ratios of 
$42\%$ and $46\%$ are much closer to the original ratios of $45\%$ among Whites (the ratio in the original data),
which we are trying to match.
Indeed, the accuracy of the second model produced by Algorithm 1 is about $2\%$ lower than the accuracy of the second model produced by Algorithm 2.
Note that some drop in accuracy is expected, since we are
measuring accuracy with respect to the original labels,
and our goal is to ensure fairness (which is not satisfied by the original labels).

%Looking at the table, we can see that the first classifier classifies subjects as frisked with a lower probability, but a clear difference between the nonwhite and white probabilities remain. After applying delta, if the second classifier is delta, the probability of being classified as frisked is further decreased and the bias is not resolved. When the second classifier is an MLP classifier, we see that the chance for a nonwhite to be frisked is brought down much closer to the white chance, showing that the bias is mostly resolved. However, we see that the AUC of the MLP classifier decreases by around 0.02.

\begin{table}
\caption{Performance of Algorithm 1 and Algorithm 2 on Stop, Question and Frisk Dataset when the first model is
of type \emph{Logistic Regression}}\label{tbl:nyc_lr}
\centering
\footnotesize
\setlength\tabcolsep{6pt}
\begin{tabular}{cc|c|c||c|c||c|c||c|c}
\cline{3-10}
& & \multicolumn{2}{c||}{ Original Data } & \multicolumn{2}{c||}{ $A^C_{(1)}$ (LR) } & \multicolumn{2}{c||}{ $C_{(2)}$ (LR) } & \multicolumn{2}{c}{ $C_{(2)}$ (MLP) } \\
\hline

\multirow{4}{*}{\rotatebox[origin=c]{90}{Algorithm 1}}\rule{0pt}{3ex} & Acc & \multicolumn{2}{c||}{} &  \multicolumn{2}{c||}{0.8186} & \multicolumn{2}{c||}{0.8181} & \multicolumn{2}{c}{0.7972} \\
\cline{2-10}

\rule{0pt}{3ex} & Filter & NA & A & NA & A & NA & A & NA & A \\ 
\cline{2-10}

%\multicolumn{1}{c|}{W} & \raggedleft {0.380} & \raggedleft {0.258} & \raggedleft {0.320} & \raggedleft {0.196} & \raggedleft {0.321} & \raggedleft {0.246}\cr
\rule{0pt}{3ex} & W & 0.456 & 0.445 & 0.321 & 0.258 & 0.320 & 0.199 & 0.323 & 0.246 \\ 
\cline{2-10}

%\multicolumn{1}{c|}{NW} & \raggedleft {0.525} & \raggedleft {0.321} & \raggedleft {0.525} & \raggedleft {0.271} & \raggedleft {0.525} & \raggedleft {0.270}\cr
\rule{0pt}{3ex} & NW & 0.588 & 0.520 & 0.525 & 0.380 & 0.525 & 0.271 & 0.525 & 0.271\\
\hline \hline

\multirow{4}{*}{\rotatebox[origin=c]{90}{Algorithm 2}}\rule{0pt}{3ex} & \multicolumn{1}{c|}{Acc} & \multicolumn{2}{c||}{} & \multicolumn{2}{c||}{0.8186} & \multicolumn{2}{c||}{0.8183} & \multicolumn{2}{c}{0.8160} \\
\cline{2-10}

\rule{0pt}{3ex} & Filter & NA & A & NA & A & NA & A & NA & A \\ 
\cline{2-10}

%\multicolumn{1}{c|}{W} & \raggedleft {0.380} & \raggedleft {0.258} & \raggedleft {0.410} & \raggedleft {0.255} & \raggedleft {0.424} & \raggedleft {0.434}\cr
\rule{0pt}{3ex} & W & 0.456 & 0.445 & 0.321 & 0.258 & 0.327 & 0.188 & 0.466 & 0.423\\
\cline{2-10}

%\multicolumn{1}{c|}{NW} & \raggedleft {0.525} & \raggedleft {0.321} & \raggedleft {0.595} & \raggedleft {0.380} & \raggedleft {0.621} & \raggedleft {0.450}\cr
\rule{0pt}{3ex} & NW & 0.588 & 0.520 & 0.525 & 0.380 & 0.528 & 0.247 & 0.604 & 0.458 \\
\hline

\end{tabular} 
\end{table}

Table~\ref{tbl:nyc_mlp} shows the performance---fairness and accuracy---for Algorithm 1 and Algorithm 2 on the \emph{Stop, Question and Frisk Dataset} when the first model, $A_{(1)}$, is of type \emph{Multi-layer Perceptron}. The first $2$ columns are obtained from the original labels of the data. They show 
that when we filter on ``Assault'', the $\mathsf{ratio}$ of individuals who got frisked
is $45\%$ and $52\%$ for White
and Non-White, respectively---a gap of
$7\%$.
The next $2$ columns are obtained
from the labels generated by
$A^C_{(1)}$ on the dataset.
In this case, when we filter on ``Assault'', the $\mathsf{ratio}$ of frisked individuals
is $25\%$ and $38\%$ for White and Non-White, respectively---a gap of $13\%$. By selecting the second model, $C_{(2)}$, to be of type \emph{Logistic Regression},
the ratios become
$23\%$ and $31\%$ for Algorithm 1
and $19\%$ and $24\%$ for Algorithm $2$.
Note that
the difference in the ratios is not significantly reduced (as expected when using a second model of type LR). Once we choose MLP as the type of the second model, $C_{(2)}$, the 
ratios become $38\%$ and $40\%$ for Algorithm 1 and
$47\%$ and $48\%$ for Algorithm $2$,
reducing the gap to $2\%$ and $1\%$, respectively.
Note that while the gap and the overall accuracy is comparable
for the two algorithms, Algorithm $2$ is preferable, since the ratios of 
$47\%$ and $48\%$ are much closer to the original ratios of $45\%$ among Whites (the ratio in the original data),
which we are trying to match.

%When MLP is used as the first classifier, the probability of being classified as frisked remains close to the original probability rather than decreasing like the Logistic Regression case. The performances of the second classifiers are similar to the previous table, with Logistic Regression being ineffective and MLP is effective in eliminating bias. In addition, AUC is not decreased in this case. Notably, the second MLP classifier actually overshoots, resulting in nonwhites being frisked less often than whites, the opposite of the original pattern. 

\begin{table}
\caption{Performance of Algorithm 1 and Algorithm 2 on Stop, Question and Frisk Dataset when the first model is
of type \emph{Multi-Layer Perceptron}}\label{tbl:nyc_mlp}
\centering
\footnotesize
\setlength\tabcolsep{6pt}
\begin{tabular}{cc|c|c||c|c||c|c||c|c}
\cline{3-10}

& & \multicolumn{2}{c||}{Original Data} & \multicolumn{2}{c||}{ $A^C_{(1)}$ (MLP) } & \multicolumn{2}{c||}{ $C_{(2)}$ (LR) } & \multicolumn{2}{c}{ $C_{(2)}$ (MLP) } \\
\hline

\multirow{4}{*}{\rotatebox[origin=c]{90}{Algorithm 1}}\rule{0pt}{3ex}   & \multicolumn{1}{c|}{Acc} & \multicolumn{2}{c||}{} & \multicolumn{2}{c||}{0.8167} & \multicolumn{2}{c||}{0.8157} & \multicolumn{2}{c}{0.8168} \\
\cline{2-10}

\rule{0pt}{3ex} & Filter & NA & A & NA & A & NA & A & NA & A \\ 
\cline{2-10}

%\multicolumn{1}{c|}{W} & \raggedleft {0.492} & \raggedleft {0.443} & \raggedleft {0.471} & \raggedleft {0.297} & \raggedleft {0.503} & \raggedleft {0.412}\cr
\rule{0pt}{3ex} & W & 0.456 & 0.445 & 0.401 & 0.412 & 0.362 & 0.230 & 0.401 & 0.378\\
\cline{2-10}

%\multicolumn{1}{c|}{NW} & \raggedleft {0.632} & \raggedleft {0.593} & \raggedleft {0.631} & \raggedleft {0.380} & \raggedleft {0.648} & \raggedleft {0.393}\cr
\rule{0pt}{3ex} & NW & 0.588 & 0.520 & 0.567 & 0.502 & 0.536 & 0.305 & 0.566 & 0.396\\
\hline \hline

\multirow{4}{*}{\rotatebox[origin=c]{90}{Algorithm 2}}\rule{0pt}{3ex} & \multicolumn{1}{c|}{Acc} & \multicolumn{2}{c||}{} & \multicolumn{2}{c||}{0.8167} & \multicolumn{2}{c||}{0.8183} & \multicolumn{2}{c}{0.8172} \\
\cline{2-10}

\rule{0pt}{3ex} & Filter & NA & A & NA & A & NA & A & NA & A \\ 
\cline{2-10}

%\multicolumn{1}{c|}{W} & \raggedleft {0.492} & \raggedleft {0.443} & \raggedleft {0.409} & \raggedleft {0.252} & \raggedleft {0.432} & \raggedleft {0.403}\cr
\rule{0pt}{3ex} & W & 0.456 & 0.445 & 0.401 & 0.412 & 0.326 & 0.188 & 0.433 & 0.473\\
\cline{2-10}

%\multicolumn{1}{c|}{NW} & \raggedleft {0.632} & \raggedleft {0.593} & \raggedleft {0.595} & \raggedleft {0.377} & \raggedleft {0.621} & \raggedleft {0.436}\cr
\rule{0pt}{3ex} & NW & 0.588 & 0.520 & 0.567 & 0.502 & 0.527 & 0.242 & 0.625 & 0.476\\
\hline
\end{tabular} 
\end{table}

\paragraph{\bf Adult Income Dataset.} 
Table~\ref{tbl:adult_lr} shows the performance---fairness and accuracy---for Algorithm 1 and Algorithm 2 on the \emph{Adult Income Dataset} when the first model, $A_{(1)}$, is of type \emph{Logistic Regression}. The first $2$ columns are obtained from the original labels of the data. They show 
that when we filter on ``Education'', the $\mathsf{ratio}$ of individuals who earn more than $50k$
is $44\%$ and $36\%$ for White
and Non-White, respectively---a gap of
$8\%$.
The next $2$ columns are obtained
from the labels generated by
$A^C_{(1)}$ on the dataset.
In this case, when we filter on ``Education'', the $\mathsf{ratio}$ of individuals who earn more than $50k$
is $62\%$ and $46\%$ for White and Non-White, respectively---a gap of $16\%$. By selecting the second model, $C_{(2)}$, to be of type \emph{Logistic Regression},
the ratios become
$63\%$ and $52\%$ for Algorithm 1
and $61\%$ and $53\%$ for Algorithm $2$.
Note that 
the difference in the ratios is only slightly reduced (as expected when using a second model of type LR). Once we choose MLP as the type of the second model, $C_{(2)}$, the 
ratios become $62\%$ and $58\%$ for Algorithm 1 and
$46\%$ and $41\%$ for Algorithm $2$,
reducing the gap to $4\%$ and $5\%$, respectively.
Note that while the gap and accuracy is comparable
for the two algorithms, Algorithm $2$ is preferable, since the ratios of 
$46\%$ and $41\%$ are much closer to the original ratios of $44\%$ among Whites (the ratio in the original data),
which we are trying to match.

%The Adult Income results are similar to the Stop and Frisk results.  Logistic Regression being ineffective in resolving the bias. MLP once again is able to mostly resolve the bias, but this time without a significant drop in AUC. 
%When using Logistic Regression as the first classifier, retraining with a second Logistic Regression classifier is unable to achieve fairness. However, using an MLP is able to nearly completely close the disparity while also maintaining a greater degree of accuracy.
%When an MLP is used as the first classifier, both Logistic Regression and MLP are able to achieve fairness, although the MLP again preserves more of the accuracy.

\begin{table}
\caption{Performance of Algorithm 1 and Algorithm 2 on Adult Income Dataset when the first model is of type \emph{Logistic Regression}}\label{tbl:adult_lr}
\centering
\footnotesize
\setlength\tabcolsep{6pt}
\begin{tabular}{cc|c|c||c|c||c|c||c|c}
\cline{3-10}

& & \multicolumn{2}{c||}{Original Data} & \multicolumn{2}{c||}{ $A^C_{(1)}$ (LR) } & \multicolumn{2}{c||}{ $C_{(2)}$ (LR) } & \multicolumn{2}{c}{ $C_{(2)}$ (MLP) } \\ 
\hline

\multirow{4}{*}{\rotatebox[origin=c]{90}{Algorithm 1}}\rule{0pt}{3ex} & \multicolumn{1}{c|}{Acc} & \multicolumn{2}{c||}{} & \multicolumn{2}{c||}{0.9005} & \multicolumn{2}{c||}{0.8917	} & \multicolumn{2}{c}{0.8998} \\ 
\cline{2-10}

\rule{0pt}{3ex} & Filter & UE & E & UE & E & UE & E & UE & E \\ 
\cline{2-10}

%\multicolumn{1}{c|}{W} & \raggedleft {0.282} & \raggedleft {0.616} & \raggedleft {0.289} & \raggedleft {0.613} & \raggedleft {0.281} & \raggedleft {0.615}\cr
\rule{0pt}{3ex} & W & 0.165 & 0.442 & 0.279 & 0.620 & 0.302 & 0.630 & 0.281 & 0.620\\
\cline{2-10}

%\multicolumn{1}{c|}{NW} & \raggedleft {0.133} & \raggedleft {0.466} & \raggedleft {0.180} & \raggedleft {0.493} & \raggedleft {0.147} & \raggedleft {0.586}\cr
\rule{0pt}{3ex} & NW & 0.088 & 0.358 & 0.143 & 0.463 & 0.195 & 0.515 & 0.160 & 0.578\\
\hline\hline

\multirow{4}{*}{\rotatebox[origin=c]{90}{Algorithm 2}}\rule{0pt}{3ex} & \multicolumn{1}{c|}{Acc} & \multicolumn{2}{c||}{} & \multicolumn{2}{c||}{0.9005} & \multicolumn{2}{c||}{0.8995	} & \multicolumn{2}{c}{0.8984} \\
\cline{2-10}

\rule{0pt}{3ex} & Filter & UE & E & UE & E & UE & E & UE & E \\ 
\cline{2-10}

%\multicolumn{1}{c|}{W} & \raggedleft {0.282} & \raggedleft {0.616} & \raggedleft {0.074} & \raggedleft {0.461} & \raggedleft {0.113} & \raggedleft {0.466}\cr
\rule{0pt}{3ex} & W & 0.165 & 0.442 & 0.279 & 0.620 & 0.260 & 0.610 & 0.100 & 0.457\\ 
\cline{2-10}

%\multicolumn{1}{c|}{NW} & \raggedleft {0.133} & \raggedleft {0.466} & \raggedleft {0.050} & \raggedleft {0.392} & \raggedleft {0.071} & \raggedleft {0.425}\cr
\rule{0pt}{3ex} & NW & 0.088 & 0.358 & 0.143 & 0.463 & 0.200 & 0.526 & 0.049 & 0.414\\
\hline

\end{tabular} 
\end{table}

Table~\ref{tbl:adult_mlp} shows the performance---fairness and accuracy---for Algorithm 1 and Algorithm 2 on \emph{Adult Income} when the first model, $A_{(1)}$, 
is of type \emph{Multi-layer Perceptron}. Similarly, the $\mathsf{ratio}$ of individuals who earn more than $50k$ is $44\%$ and $27\%$ for White and Non-White, respectively. Using a second model of type \emph{Logistic Regression} reduces the difference to $5\%$. Using a second model of type \emph{Multi-layer Perceptron} reduces the difference even further to $2\%$ for Algorithm 2. There is also no significant drop in terms of accuracy of the classifier.

Table~\ref{tbl:adult_mlp} shows the performance---fairness and accuracy---for Algorithm 1 and Algorithm 2 on the \emph{Adult Income Dataset} when the first model, $A_{(1)}$, is of type \emph{Multi-layer Perceptron}. The first $2$ columns are obtained from the original labels of the data. They show 
that when we filter on ``Assault'', the $\mathsf{ratio}$ of individuals who earn more than $50k$
is $44\%$ and $36\%$ for White
and Non-White, respectively---a gap of
$8\%$.
The next $2$ columns are obtained
from the labels generated by
$A^C_{(1)}$ on the dataset.
In this case, when we filter on ``Education'', the $\mathsf{ratio}$ 
of individuals who earn more than $50k$
is $44\%$ and $28\%$ for White and Non-White, respectively---a gap of $16\%$. By selecting the second model, $C_{(2)}$, to be of type \emph{Logistic Regression},
the ratios become
$55\%$ and $50\%$ for Algorithm 1
and $61\%$ and $53\%$ for Algorithm $2$.
Note that
the difference in the ratios is only slightly reduced in the case of Algorithm $2$ (as expected when using a second model of type LR). Once we choose MLP as the type of the second model, $C_{(2)}$, the 
ratios become $46\%$ and $40\%$ for Algorithm 1 and
$44\%$ and $41\%$ for Algorithm $2$,
reducing the gap to $6\%$ and $3\%$, respectively.
Note that while the gap and the overall accuracy is comparable
for the two algorithms, Algorithm $2$ is preferable, since the ratios of 
$44\%$ and $41\%$ are much closer to the original ratios of $44\%$ among Whites (the ratio in the original data),
which we are trying to match.

\begin{table}
\caption{Performance of Algorithm 1 and Algorithm 2 on Adult Income Dataset when the first model is of type
\emph{Multi-Layer Perceptron}}\label{tbl:adult_mlp}
\centering
\footnotesize
\setlength\tabcolsep{6pt}
\begin{tabular}{cc|c|c||c|c||c|c||c|c}
\cline{3-10}

& & \multicolumn{2}{c||}{Original Data} & \multicolumn{2}{c||}{ $A^C_{(1)}$ (MLP) } & \multicolumn{2}{c||}{ $C_{(2)}$ (LR) } & \multicolumn{2}{c}{ $C_{(2)}$ (MLP) } \\ 
\hline

\multirow{4}{*}{\rotatebox[origin=c]{90}{Algorithm 1}}\rule{0pt}{3ex} & \multicolumn{1}{c|}{Acc} & \multicolumn{2}{c||}{} & \multicolumn{2}{c||}{0.8994} & \multicolumn{2}{c||}{0.8927} & \multicolumn{2}{c}{0.9036} \\ 
\cline{2-10}

\rule{0pt}{3ex} & Filter & UE & E & UE & E & UE & E & UE & E \\ 
\cline{2-10}

%\multicolumn{1}{c|}{W} & \raggedleft {0.102} & \raggedleft {0.475} & \raggedleft {0.069} & \raggedleft {0.475} & \raggedleft {0.100} & \raggedleft {0.474}\cr
\rule{0pt}{3ex} & W & 0.165 & 0.442 & 0.086 & 0.442 & 0.131 & 0.549 & 0.087 & 0.457\\
\cline{2-10}

%\multicolumn{1}{c|}{NW} & \raggedleft {0.076} & \raggedleft {0.284} & \raggedleft {0.067} & \raggedleft {0.429} & \raggedleft {0.078} & \raggedleft {0.444}\cr
\rule{0pt}{3ex} & NW & 0.088 & 0.358 & 0.071 & 0.276 & 0.118 & 0.500 & 0.083 & 0.403\\
\hline \hline

\multirow{4}{*}{\rotatebox[origin=c]{90}{Algorithm 2}}\rule{0pt}{3ex} & \multicolumn{1}{c|}{Acc} & \multicolumn{2}{c||}{} & \multicolumn{2}{c||}{0.8994} & \multicolumn{2}{c||}{0.8991} & \multicolumn{2}{c}{0.8982} \\ 
\cline{2-10}

\rule{0pt}{3ex} & Filter & UE & E & UE & E & UE & E & UE & E \\ 
\cline{2-10}

%\multicolumn{1}{c|}{W} & \raggedleft {0.102} & \raggedleft {0.475} & \raggedleft {0.074} & \raggedleft {0.464} & \raggedleft {0.108} & \raggedleft {0.462}\cr
\rule{0pt}{3ex}  & W & 0.165 & 0.442 & 0.086 & 0.442 & 0.259 & 0.605 & 0.095 & 0.435\\ 
\cline{2-10}

%\multicolumn{1}{c|}{NW} & \raggedleft {0.076} & \raggedleft {0.284} & \raggedleft {0.053} & \raggedleft {0.403} & \raggedleft {0.068} & \raggedleft {0.410}\cr
\rule{0pt}{3ex}  & NW & 0.088 & 0.358 & 0.071 & 0.276 & 0.203 & 0.526 & 0.047 & 0.410\\
\hline

\end{tabular} 
\end{table}

\section{Fairness via Classification Parity}\label{sec:classification_parity}
\begin{figure}[ht]
\begin{minipage}{0.49\linewidth}
\centerline{\includegraphics[width=1\linewidth]{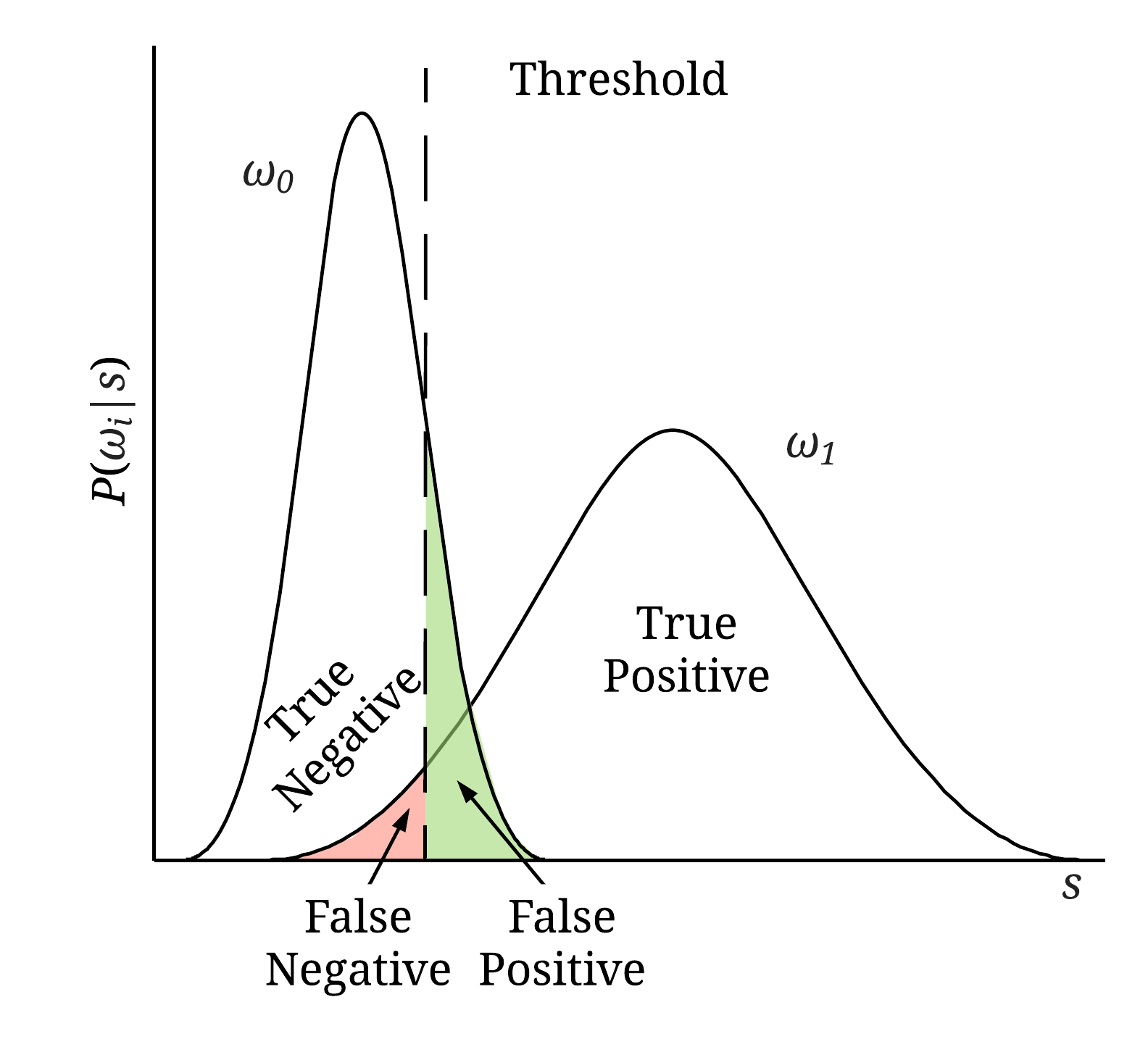}}
\centerline{(a)}
\end{minipage}
\hfill
\begin{minipage}{0.49\linewidth}
\centerline{\includegraphics[width=1\linewidth]{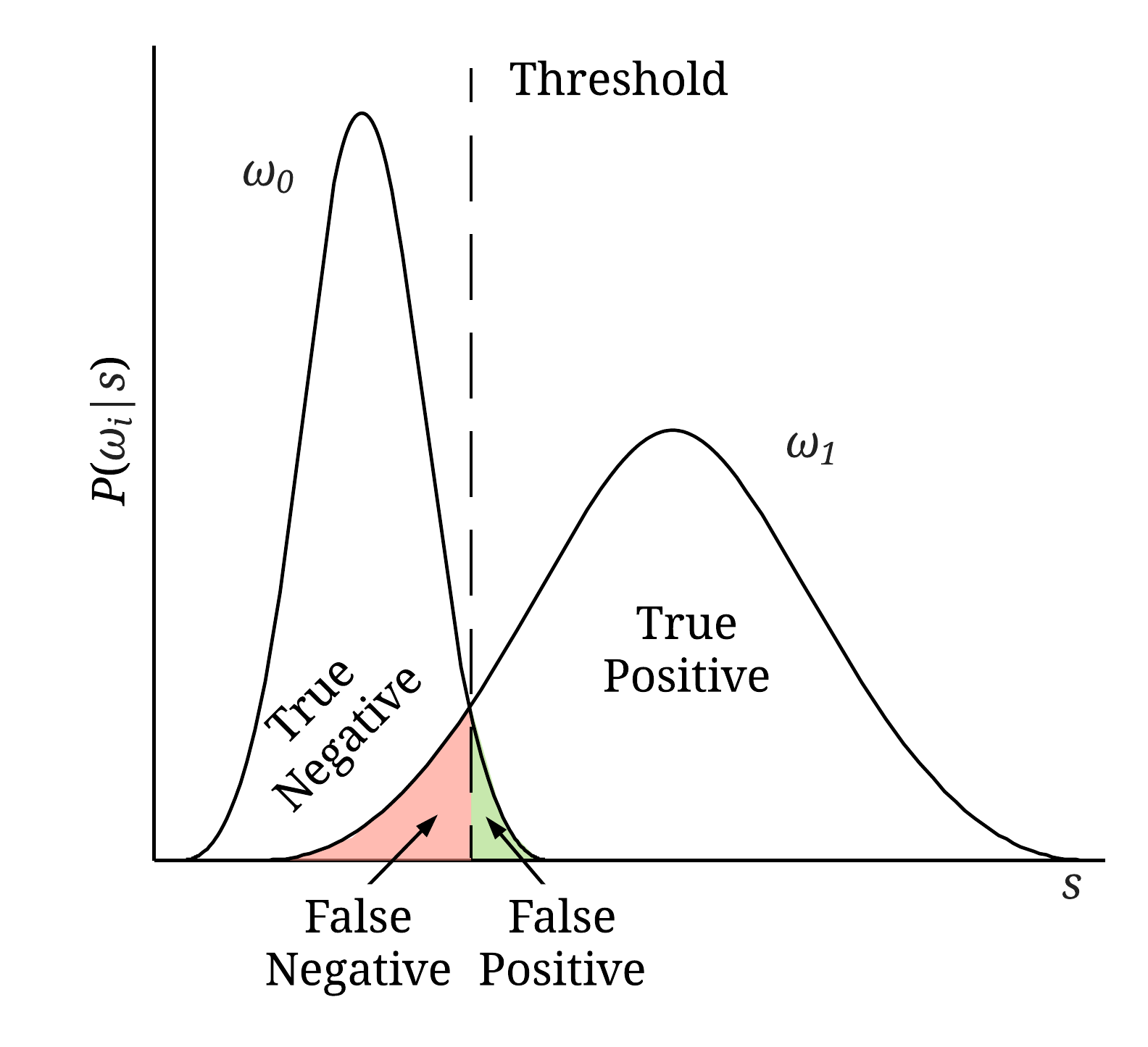}}
\centerline{(b)}
\end{minipage}
\caption{The posterior distribution and the decision threshold. (a) Relations of TP and FP with the given threshold. (b) The threshold of the minimum error classifier.}
\label{fig:score_distr}
\end{figure}

In this section, we investigate the fairness issue from statistical learning perspective. In Bayesian decision theory for binary classification, one sample $s$ is classified into negative class $\omega_0$ or positive $\omega_1$ by one decision threshold, which is determined according to the posterior probability distribution. For example, given the distribution and the threshold shown in Fig.~\ref{fig:score_distr}(a), we have the relation of true positive (TP), true negative (TN), false positive (FP) and false negative (FN). Specially for the minimum error classifier, the decision threshold is set as Fig.~\ref{fig:score_distr}(b), where $P(\omega_0|s)=P(\omega_1|s)$. When the decision threshold is tuned, the true positive rate (TPR) and false positive rate (FPR) are altered accordingly. 

In general, a binary classifier $C$ can be considered as a mapping function of the input sample $x$ to the posterior probability or score $s$, i.e., $C:x\rightarrow s$. Denote $y$ and $\hat{y}$ are the corresponding label and classifier prediction of $x$, and $y,\hat{y}\in\{0,1\}$. The classifier's output scores from the group of data formulates the score distribution shown in Fig.~\ref{fig:score_distr}. The decision criterion is

\begin{equation}
    \hat{y}=\left\{\begin{matrix}
        0,~~~s=C(x)<t\\ 
        1,~~~s=C(x)\geq t
    \end{matrix}\right.
\end{equation}
where $t$ is the decision threshold. When we tune the decision threshold for one trained classifier $C$, we can obtain different TPR and FPR according to the score distribution, which formulates $C$'s ROC curve.

%Receiver operating characteristic 

\subsection{Classification Parity}

Suppose the dataset $D$ are separated into multiple completely exhaustive mutually exclusive groups $D_1, D_2,..., D_K$ based on the protected attribute. For example, if one classifier has higher TPR and FPR in the group $D_k$ than the other group $D_i, i\neq k$, the classifier is prone to provide positive inferences in $D_k$. We refer to the above condition as \emph{positively biased} in $D_k$. Prior art of the fairness definition, such as demographic parity~\cite{zafar2017demographic}, equalized odds~\cite{hardt2016equality} and predictive rate parity~\cite{zafar2017fairness}, can alleviate the biased prediction in the classifier by equalizing the performance statistics of the classifier among all the groups. However, these classification parity conditions guarantees that the classifier only satisfies the parity condition at one specific threshold setting. If we modify the decision threshold, the classification parity condition no longer holds in the classifier performance.

From Bayesian decision theory perspective, one fundamental reason of the bias predictions in ML models comes from the intrinsic disparities of the score distributions from the given classifier among all the groups. To achieve fairness in ML, one feasible solution is to alleviate the discrepancy of the score distributions in different groups. Once the distributions are equalized, the classifier performances are identical and indistinguishable among all the groups, whatever the decision threshold is. We propose the definition of equalized distribution as follows.

%\begin{definition} [Classification parity] A binary classifier $C$ achieves weak classification parity, if TPR and FPR among the groups $D_k, k=1,2,...,K$ are identical, i.e.,
%    \begin{equation}
%        \begin{matrix}
%        P(\hat{y}=1|z\in D_i, y=1)=P(\hat{y}=1|z\in D_j, y=1), \forall i,j~~(TPR~equality)\\
%        P(\hat{y}=1|z\in D_1, y=0)=P(\hat{y}=1|z\in D_2, y=0), \forall i,j~~(FPR~equality)
%        \end{matrix}
%    \end{equation}
%This condition is named as equalized odds~\cite{hardt2016equality}.
\begin{definition} [Equalized distribution] A binary classifier $C$ achieves equalized distribution, if the score distributions are identical among all the groups $D_k, k=1,2,...,K$, i.e.,
\begin{equation}
    \textup{pdf}(C(D_i))=\textup{pdf}(C(D_j)),~\forall i,j=1,2,...,K
\end{equation}
where \textup{pdf} stands for probability density function, which computes the data distribution.
\end{definition}

More strictly, equalized distribution enforces the equalization of the group-wise score distributions from the classifier. One implication is that equalized distribution makes the prior art of the classification parity, such as demographic parity, equalized odds, always holds true whatever the decision threshold is.
Hence, unlike the prior art of the parity notions, ``equalized distribution'' is \emph{threshold-invariant}. The equalized distribution requires the classifier to offer equivalent and indistinguishable learning performance statistics among all the groups $D_k, k=1,2,...,K$ independent of the decision threshold. To contrast the notion of the prior art of the classification parity and the proposed equalized distribution, we refer to the prior art as the weak condition of classification parity, and equalized distribution as the strong condition.

Based on the classification parity, we propose the fairness ML algorithms in two possible scenarios: with and without prior knowledge of the protected attribute in test stage.

\subsection{With Prior Knowledge of the Protected Attribute}
\label{sec:with_knowledge}

In this scenario, we assume that we have the prior knowledge of the protected attribute of the input samples in the test stage. We enforce equalized odds (i.e., weak classification parity) on the classifier. Given a trained classifier $C$, the condition of equalized odds can be satisfied by tuning the decision thresholds for different groups. We denote the number of correct predictions in group $D_k, k=1,2,...,K$ with the decision threshold $t_k$ as $N(D_k,t_k)$. Then the classification accuracy can be written as
\begin{equation}
    E_a = \frac{\sum_{k=1}^{K}N(D_k,t_k)}{\sum_{k=1}^K|D_k|}
\end{equation}
To achieve the equalized odds, we tune the thresholds $t_k, k=1,2,...,K$ to shrink the discrepancy of TPR and FPR among all the groups, which can be represented as the fairness term
\begin{equation}
    E_f = \sum_{k=2}^K\Big(|\mathrm{TPR}(D_1,t_1)-\mathrm{TPR}(D_k,t_k)|+|\mathrm{FPR}(D_1,t_1)-\mathrm{FPR}(D_k,t_k)|\Big)
\end{equation}
where $\mathrm{TPR}(D_k,t_k)$ and $\mathrm{FPR}(D_k,t_k)$ denote true positive rate and false positive rate in group $D_k$ with the decision threshold $t_k$. In practice, we trade-off the classification accuracy and the fairness term. Hence, given the trained classifier $C$, the fairness problem becomes an optimization problem
\begin{equation}
\label{eqn:with_knowledge}
    \max_{t_k,k=1,2,...,K}E_a-\lambda E_f
\end{equation}
where $\lambda$ is a hyperparameter. Since Eqn.~\ref{eqn:with_knowledge} is a nonlinear and non-differentiable problem, we apply particle swarm optimization (PSO)~\cite{kennedy2010particle} to find the best set of thresholds $\{t_k\}_{k=1,2,...K}$ for $K$ groups in terms of reducing the difference of TPR and FPR between two groups.

\subsection{Without Prior Knowledge of the Protected Attribute}

In this scenario, we assume that no prior knowledge of the protected attribute of the input samples is provided in the test stage. Different from Section~\ref{sec:with_knowledge}, we enforce strong classification parity on the trained classifier. The equalized distribution condition sufficiently guarantees that the classifier can achieve classification parity among all the groups using one universal and non-specific threshold. 

How can we formulate the constraint in terms of equalized distribution? We propose to approximate the continuous score distributions using the discrete histograms. Denote the sample $x_i$ and the corresponding decision score $s_i$ provided by the given classifier $C$, i.e., $s_i = C(x_i)$. The count of the scores in the bin $(c-\frac{\Delta}{2},c+\frac{\Delta}{2})$ can be expressed as
\begin{equation}
    n_c=\sum_{i}rect_c(s_i)
\end{equation}
where $c$, $\Delta$ are the center and the bandwidth of the bin, and $rect_c(\cdot)$ is a rectangular function

\begin{equation}
    rect_c(s)=
    \left\{\begin{matrix}
1,&s\in(c-\frac{\Delta}{2},c+\frac{\Delta}{2})\\ 
0,&o.w
\end{matrix}\right.
\end{equation}

Since such function is not differentiable, we approximate it with Gaussion function, i.e.,

\begin{equation}
    n_c=\sum_{i}gauss_c(s_i)=\sum_{i}\exp\big(-\frac{(s_i-c)^2}{2\sigma^2}\big)
\end{equation}

Thus, the histogram of the decision score in group $D_k$ can be expressed as

\begin{equation}
    hist(D_k)=\sum_{c}\sum_{x\in D_k}gauss_c(C(x))
\end{equation}
We normalize the histogram as
\begin{equation}
    norm\_hist(D_k)=\frac{1}{|D_k|}hist(D_k)
\end{equation}

We use $L$-2 loss to constraint the equalization of the score distributions in two groups, i.e.
\begin{equation}
\begin{aligned}
    E_f=&\sum_{k=2}^K\Big(||norm\_hist(D_{1,p})-norm\_hist(D_{k,p})||_2^2\\
    &~~+||norm\_hist(D_{1,n})-norm\_hist(D_{k,n})||_2^2\Big)
\end{aligned}
\end{equation}
where $D_{k,p}$ denotes the set of positive samples in group $D_k$ and $D_{k,n}$ denotes the set of negative samples in $D_k$. Like Section~\ref{sec:with_knowledge}, we formulate the minimization problem, trade-off the classification accuracy and the fairness term.
\begin{equation}
    L=\alpha E_{a}+(1-\alpha)E_f
\label{eqn:without_knowledge}
\end{equation}
where $E_{a}$ is the accuracy term, such as mean square error (MSE) or logistic loss, and $\alpha \in [0,1]$. As the loss function $L$ is continuous, gradient based methods can be employed to search the solution.

\subsection{Experimental Results}
We conducted the experiments on the COMPAS dataset~\cite{dressel2018accuracy}. The COMPAS records the recidivation of 7214 criminals in total. Each line of record documents the criminal's information of race, sex, age, number of juvenile felony criminal charges (juv\_fel\_count), number of juvenile misdemeanor criminal charges (juv\_misd\_count), number of non-juvenile criminal charges (priors\_count), the previous criminal charge (charge\_id), the degree of the charge (charge\_degree) and the ground truth of two-year recidivation (two\_year\_recid). In the experiments, we chose race as the protected attribute and, for simplicity, only focused on two race groups, ``white'' and ``black'', excluding the records of other races. Hence, we totally used 6150 samples to demonstrate the effectiveness of the proposed fairness ML algorithms in two groups. In the experiments, we randomly splited the dataset into $60\%$ training, $20\%$ validation, and $20\%$ test.
\subsubsection{Feature Preprocessing} Among the unprotected features of one criminal record, binary features, i.e., sex and charge\_degree, are employed 0-1 encoding. Continuous features, i.e., age, juv\_fel\_count, juv\_misd\_count and priors\_count are applied with data standardization. The categorical feature, charge\_id, is encoded with one-hot encoding, leading to a 430-D feature vector. Thus, the total dimension of the feature vector is 436. Then, we applied principle component analysis (PCA) to reduce the 436-D feature vector to 20-D feature vector.

According to the protected feature (e.g. race), the dataset $D$ is divided into the ``white'' group $D_w$ and the ``black'' group $D_b$. Our goal is to obtain a binary classifier to offer fair predictions of the two-year recidivation in $D_w$ and $D_b$.
\subsubsection{With Prior Knowledge}
We investigated two machine learning models in this scenario: logistic regression and SVM. We first trained the model with the training set as normal, and then tuned the decision thresholds on the validation set. The finalized classifier was tested on the test set. During the tuning, we set $\lambda = 1$ in Eqn.~\ref{eqn:with_knowledge} to trade-off the accuracy and the classification parity. 

\begin{figure}[ht]
\begin{minipage}{0.49\linewidth}
\centerline{\includegraphics[width=1\linewidth]{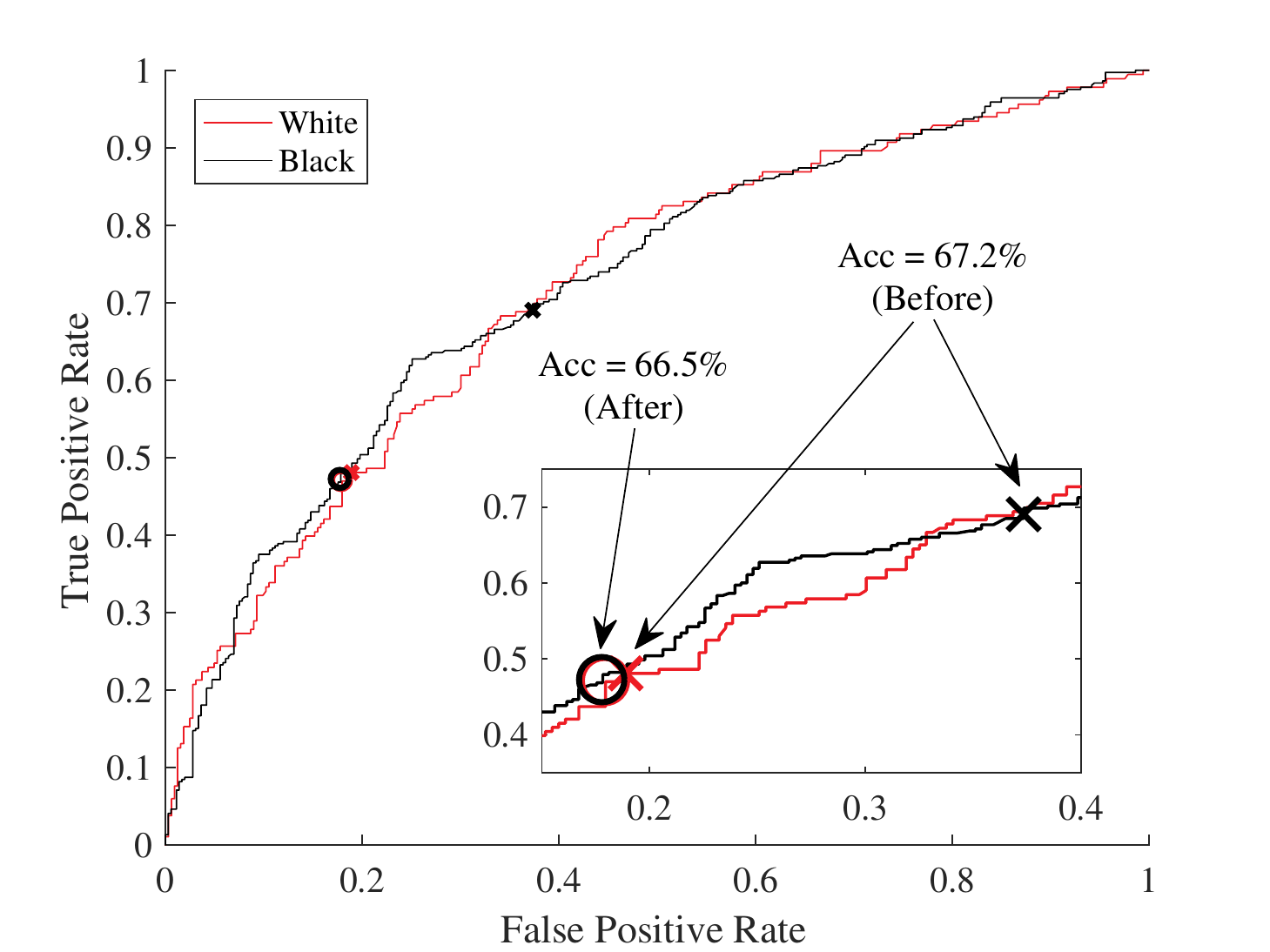}}
\centerline{(a)}
\end{minipage}
\hfill
\begin{minipage}{0.49\linewidth}
\centerline{\includegraphics[width=1\linewidth]{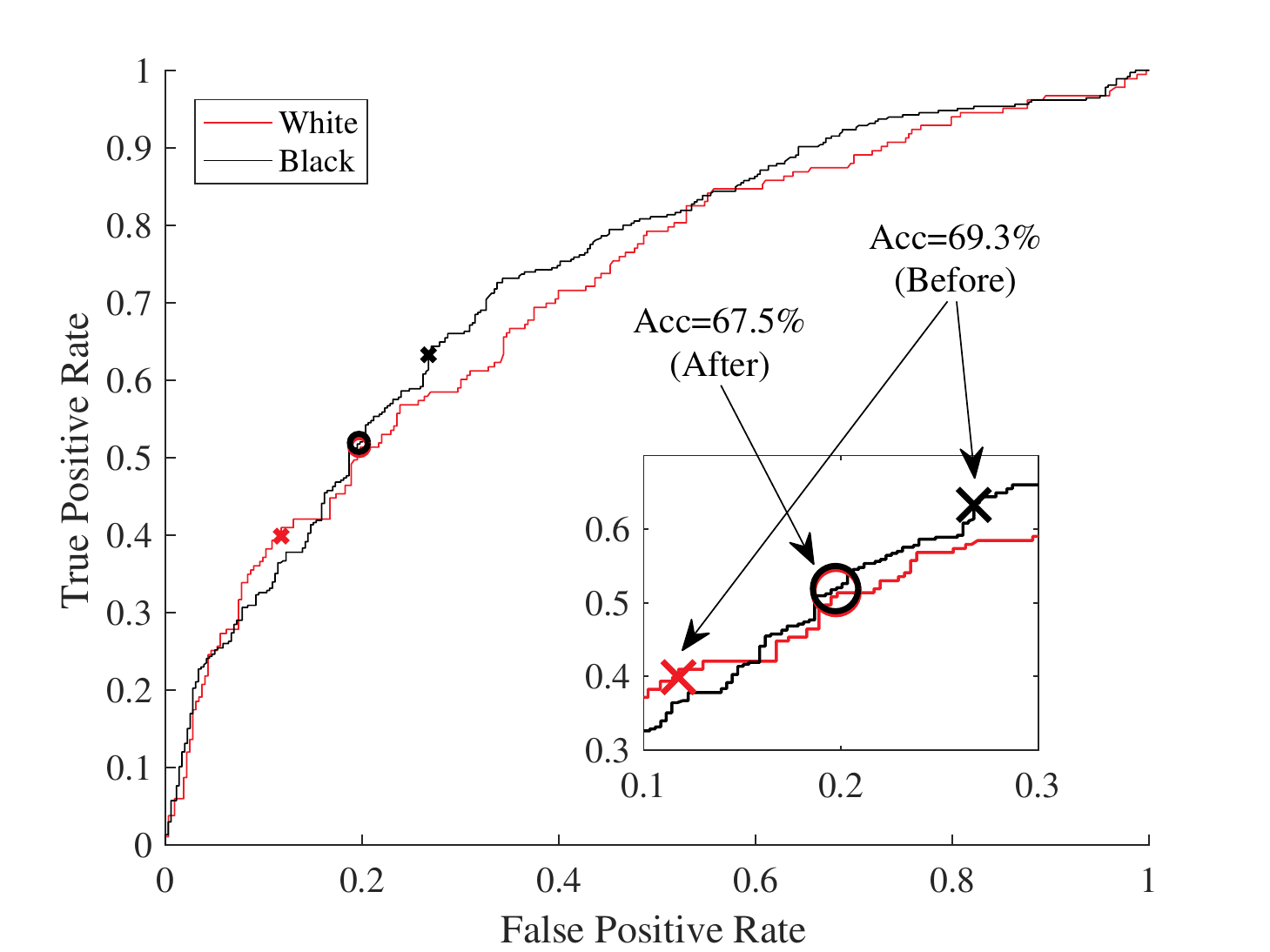}}
\centerline{(b)}
\end{minipage}
\caption{The state of the classifier on the ROC curves on the validation set. The Crosses represent the states before tuning, and circles represent the states after tuning. (a) Logistic regression classifier. (b) SVM classifier.}
\label{fig:tuning_roc}
\end{figure}
Table~\ref{tab:tuning} presents the performance of the classifiers before and after tuning. Before tuning, the logistic regression classifier has the default decision threshold $0.5$, while the SVM classifier has the default decision threshold $0$. According to the statistics of TPR and FPR before tuning, we can see both of the classifiers are positively biased in the group ``black'' to ``white''. The tuning algorithm is employed on the validation set to shrink the discrepancy between two groups. Fig.~\ref{fig:tuning_roc} shows the state of the classifier on the ROC curves before and after tuning on the validation set. We can see that the tuning algorithm help two points merge to an overlap point on the curves to fulfill the classification parity. After the tuning algorithm on the validation set, the gaps of TPR and FPR on test set between two groups were successfully reduced from around $0.2$ to no more than $0.05$, but the drop of the overall accuracy is insignificant. Specifically, the logistic regression classifier only decreases $1.7\%$ on classification accuracy and the SVM classifier lowers the accuracy by $2.3\%$.

\begin{table}[ht]
\centering
\begin{threeparttable}[b]
\caption{The performance of the classifiers before and after tuning}
\begin{tabular}{cc|c|c|c|c||cccc}
\hline
\multicolumn{2}{c|}{\multirow{2}{*}{Tuning}} & \multicolumn{4}{c||}{Logistic}                        & \multicolumn{4}{c}{SVM}                                                                                             \\ \cline{3-10} 
\multicolumn{2}{c|}{}                        & THR & Acc                & TPR   & FPR   & \multicolumn{1}{c|}{THR} & \multicolumn{1}{c|}{Acc}                & \multicolumn{1}{c|}{TPR}   & FPR   \\ \hline
\multirow{2}{*}{~Before~}        & ~White~       & 0.5        & \multirow{2}{*}{67.9\%} & 0.407 & 0.178 & \multicolumn{1}{c|}{0}          & \multicolumn{1}{c|}{\multirow{2}{*}{67.6\%}} & \multicolumn{1}{c|}{0.346} & 0.143 \\ \cline{2-3} \cline{5-7} \cline{9-10} 
                               & ~Black~       & 0.5        &                         & 0.716 & 0.337 & \multicolumn{1}{c|}{0}          & \multicolumn{1}{c|}{}                        & \multicolumn{1}{c|}{0.620} & 0.252 \\ \hline
\multirow{2}{*}{~After~}         & ~White~       & 0.503      & \multirow{2}{*}{~66.2\%~} & 0.401 & 0.160 & \multicolumn{1}{c|}{-0.281}     & \multicolumn{1}{c|}{\multirow{2}{*}{~65.3\%~}} & \multicolumn{1}{c|}{0.456} & 0.205 \\ \cline{2-3} \cline{5-7} \cline{9-10} 
                               & ~Black~       & ~~0.588~~      &                         & ~~0.445~~ & ~~0.147~~ & \multicolumn{1}{c|}{~~0.300~~}      & \multicolumn{1}{c|}{}                        & \multicolumn{1}{c|}{~0.478~~} & ~~0.198~~ \\ \hline
\end{tabular}
\begin{tablenotes}
\scriptsize
 \item $^1$THR: Decision threshold~~~~~~~~~~~~~~~~~~~~~~~~~~$^2$Acc: Total classification accuracy
 \item $^3$TPR: True positive rate~~~~~~~~~~~~~~~~~~~~~~~~~~~$^4$FPR: False positive rate
\end{tablenotes}
\label{tab:tuning}
\end{threeparttable}
\end{table}

To sum up, the proposed tuning method enforces the fairness by tuning the decision thresholds of one given model for all the groups. It can be applied after any trained classifiers without modified them. On the other hand, the main disadvantage is that it requires access to the protected attribute of the samples in test stage. Although the tuning method can, to some extent, alleviates the biased prediction in ML model, the model itself still has distinguishable learning performances in two groups in terms of the score distributions.

\subsubsection{Without Prior Knowledge} In this scenario, we conducted experiments with logistic regression classifier. The accuracy term $E_a$ in Eqn.~\ref{eqn:without_knowledge} is the logistic loss for binary classification, i.e.,
\begin{equation}
    E_a = \sum_i y_i\log C(x_i)+(1-y_i)\log (1-C(x_i))
\end{equation}
where $C$ is the trained classifier, and $(x_i, y_i)$ is the sample and label pair in the whole dataset $D$. We set the bandwidth of the bin in the histogram $\Delta$ as 0.02. Since the output of the logistic regression classifier is confined in the range $[0,1]$, the center of the bins are 0.01, 0,03, 0.05,..., 0.97 and 0.99. In the training, we applied momentum gradient descent to optimize the loss function in Eqn.~\ref{eqn:without_knowledge}, with learning rate $\mu=1$ and momentum $=0.9$. The number of learning iteration is $2$k.

\begin{figure}[ht]
    \begin{minipage}{0.49\linewidth}
      \centerline{\includegraphics[width=1.0\linewidth]{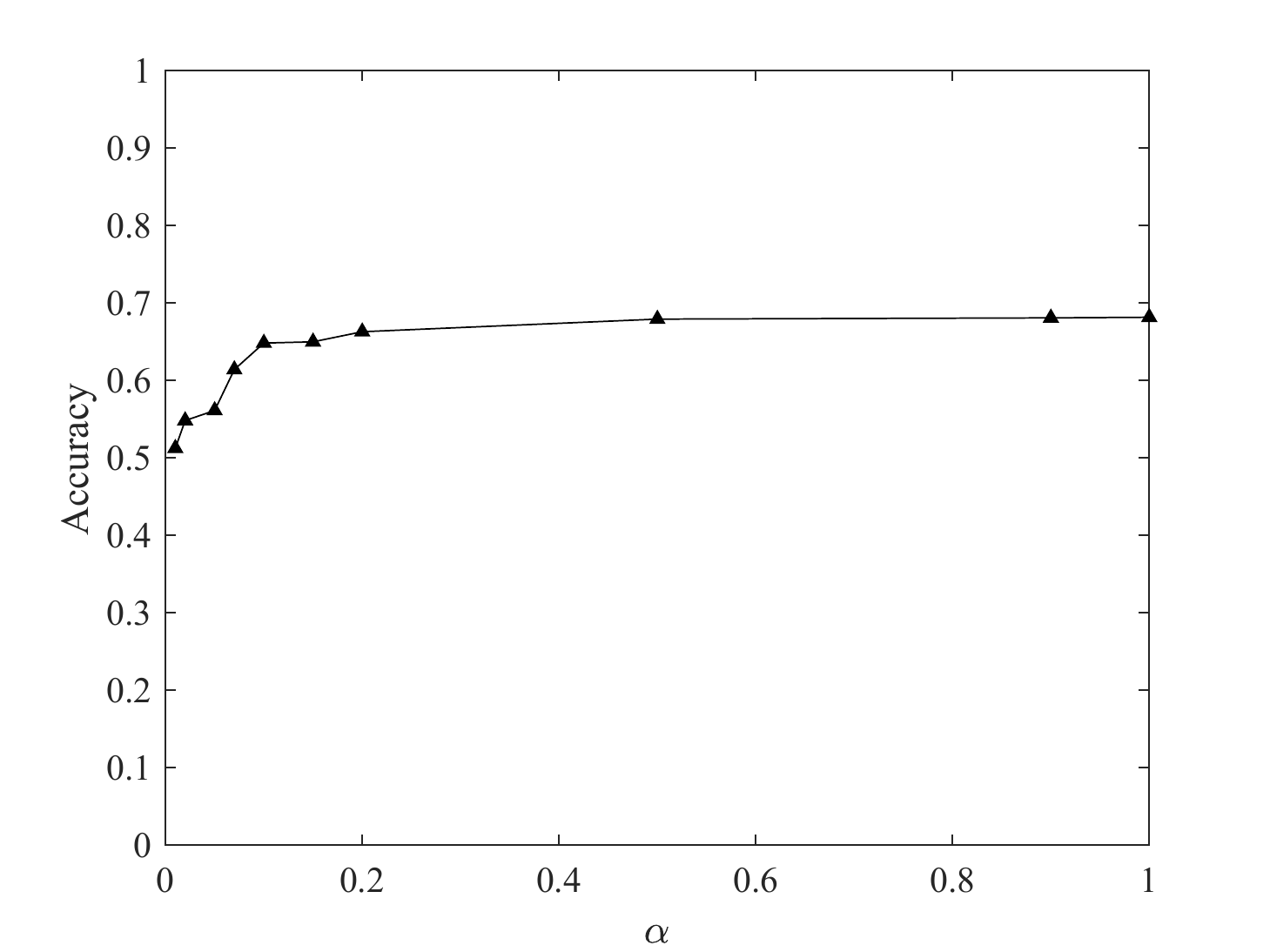}}
      \centerline{(a)}
    \end{minipage}
    \hfill
    \begin{minipage}{.49\linewidth}
      \centerline{\includegraphics[width=1.0\linewidth]{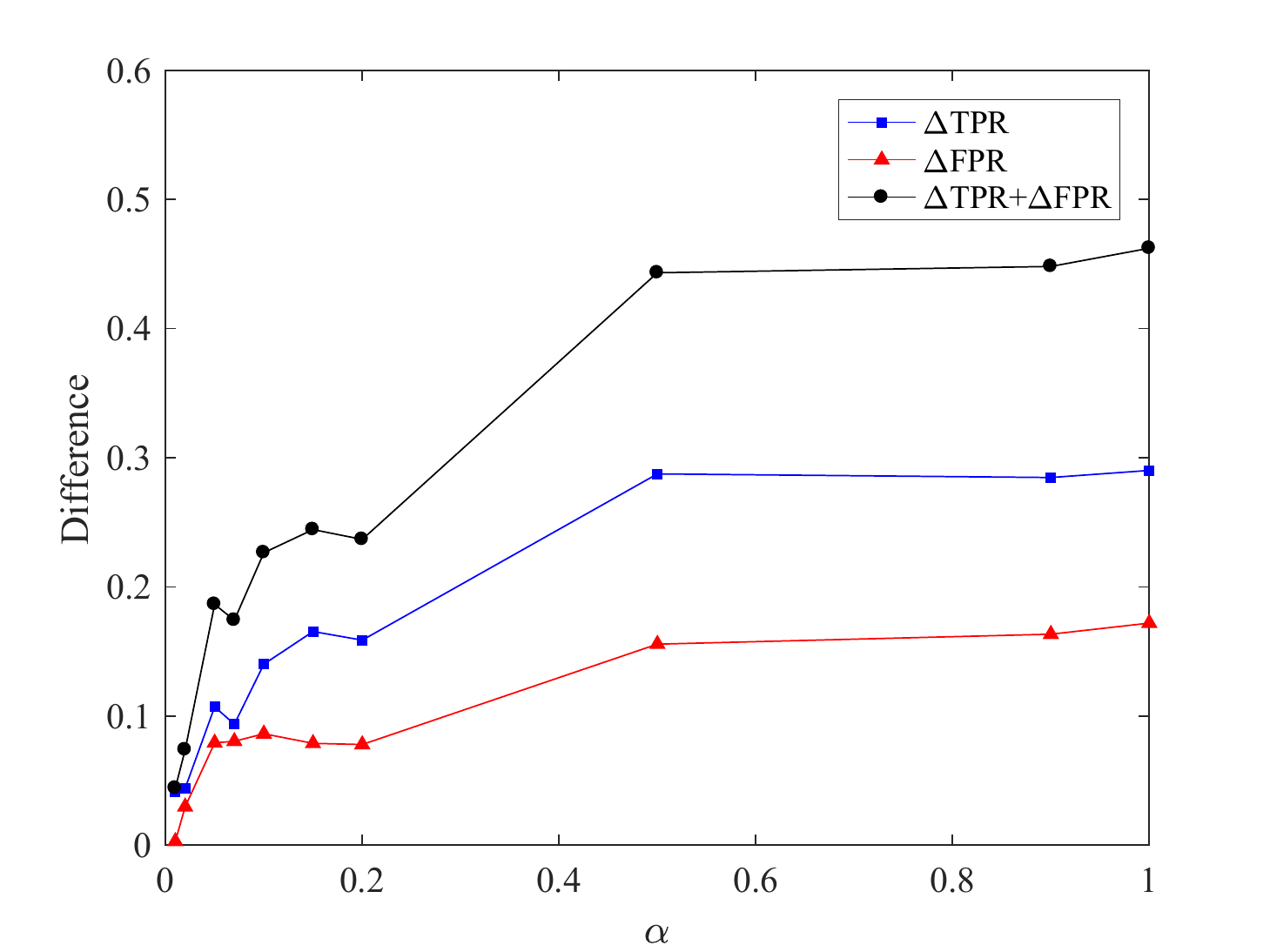}}
      \centerline{(b)}
    \end{minipage}
    \caption{The influence of the hyperparameter $\alpha$ on the performance of the classifier. (a) The overall accuracy of the classifier with different $\alpha$. (b) The fairness indicator of the classifier with different $\alpha$, i.e., the difference of TPR/FPR between two groups.}
    \label{fig:alpha}
\end{figure}

\begin{figure}[ht]
%\begin{tabular}{cc}   
\begin{minipage}{0.495\linewidth}
      \centerline{\includegraphics[width=1.0\linewidth]{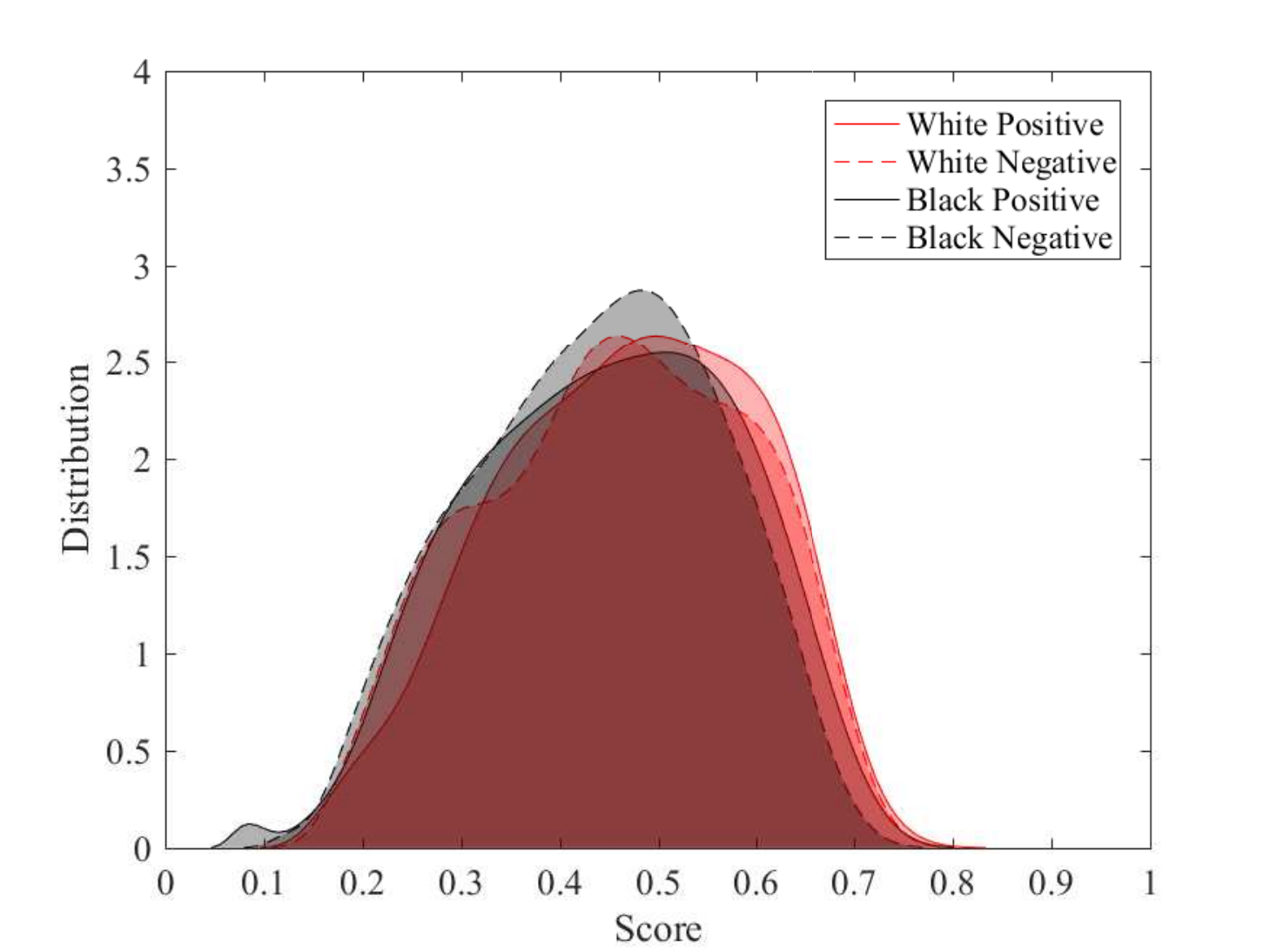}}
  \centerline{(a)}
\end{minipage}
\hfill
\begin{minipage}{0.495\linewidth}
      \centerline{\includegraphics[width=1.0\linewidth]{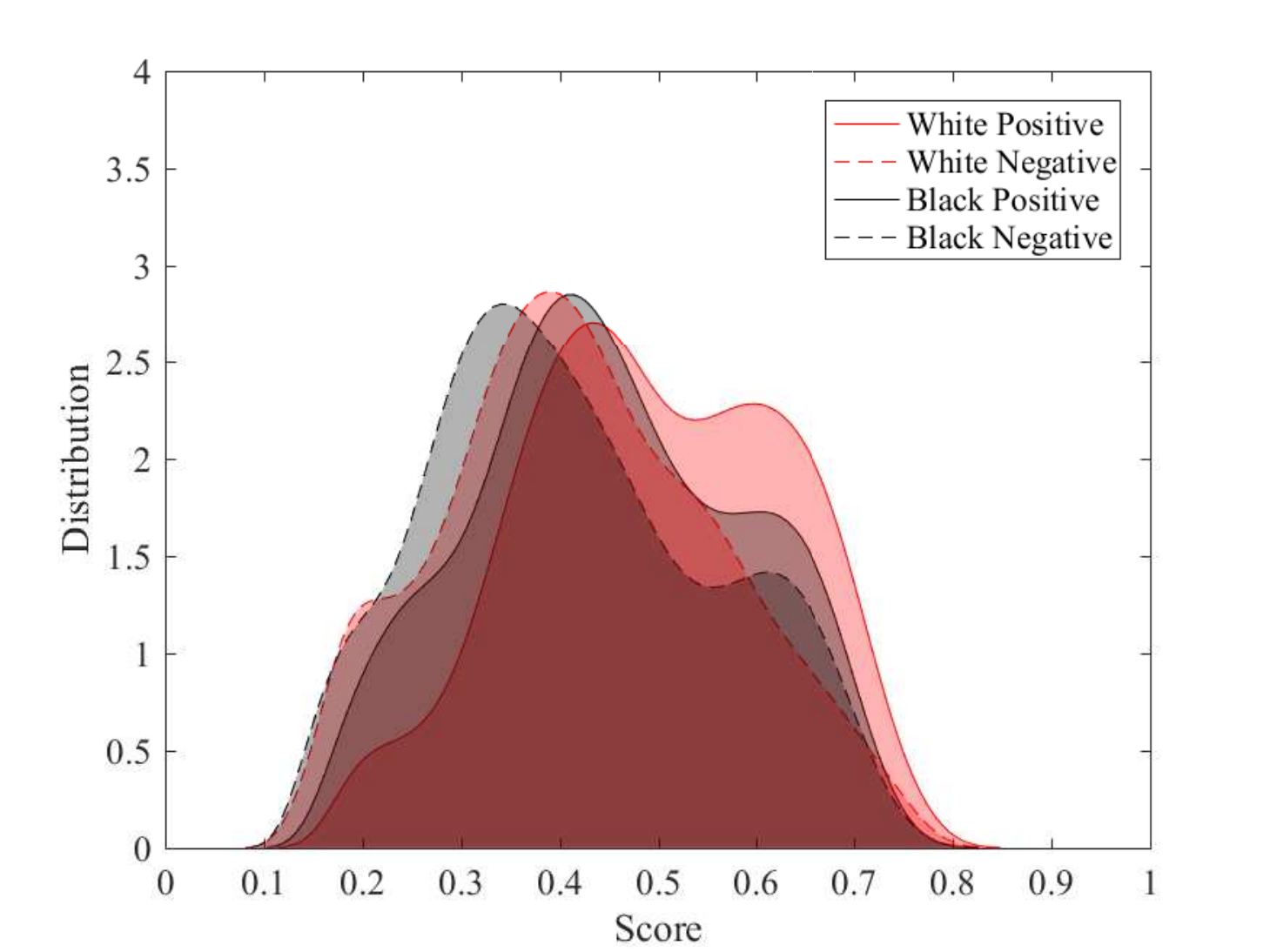}}
      \centerline{(b)}
\end{minipage}
\vfill
\begin{minipage}{0.495\linewidth}
      \centerline{\includegraphics[width=1.0\linewidth]{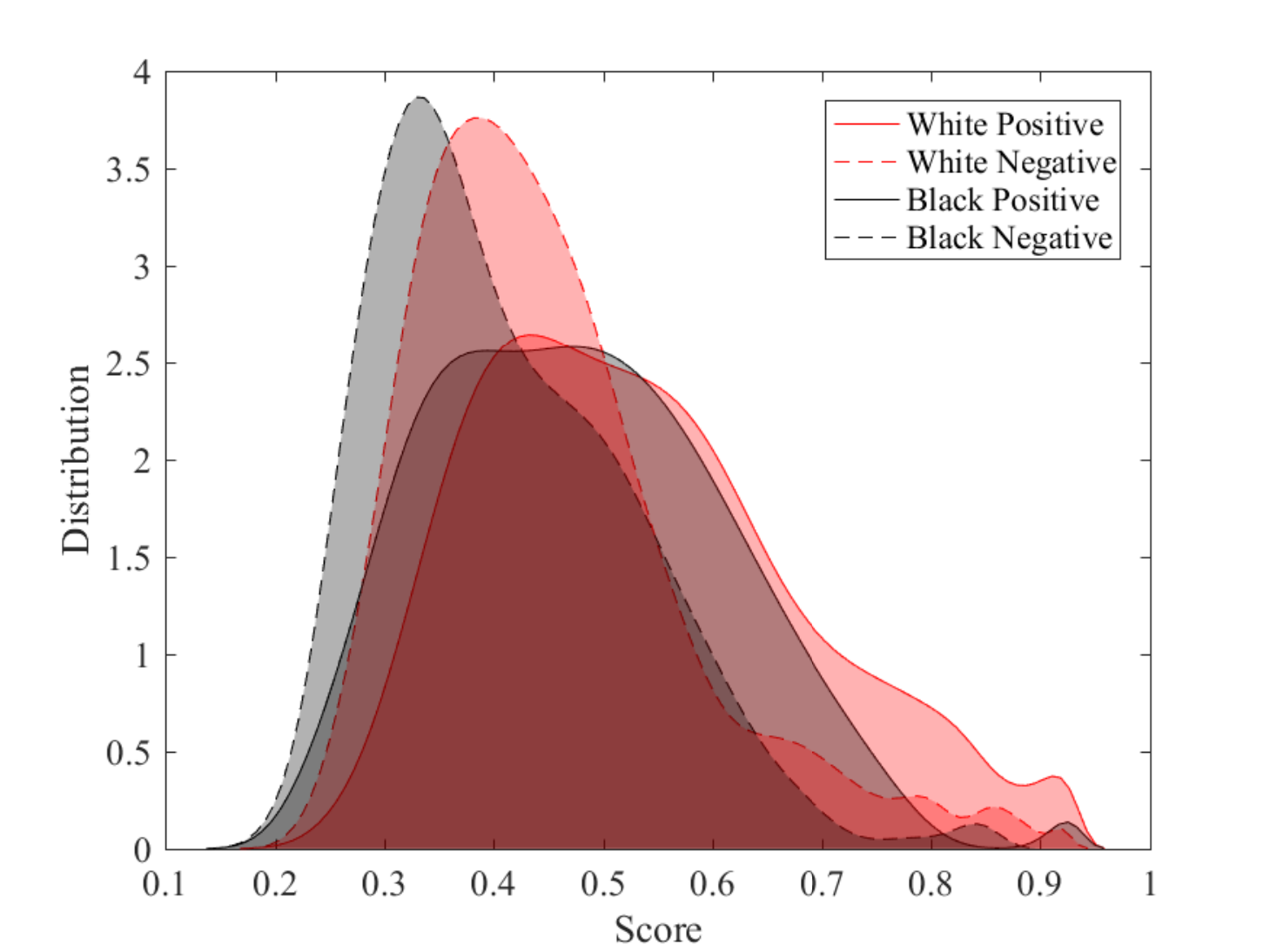}}
      \centerline{(c)}
\end{minipage}
\hfill
\begin{minipage}{0.495\linewidth}
      \centerline{\includegraphics[width=1.0\linewidth]{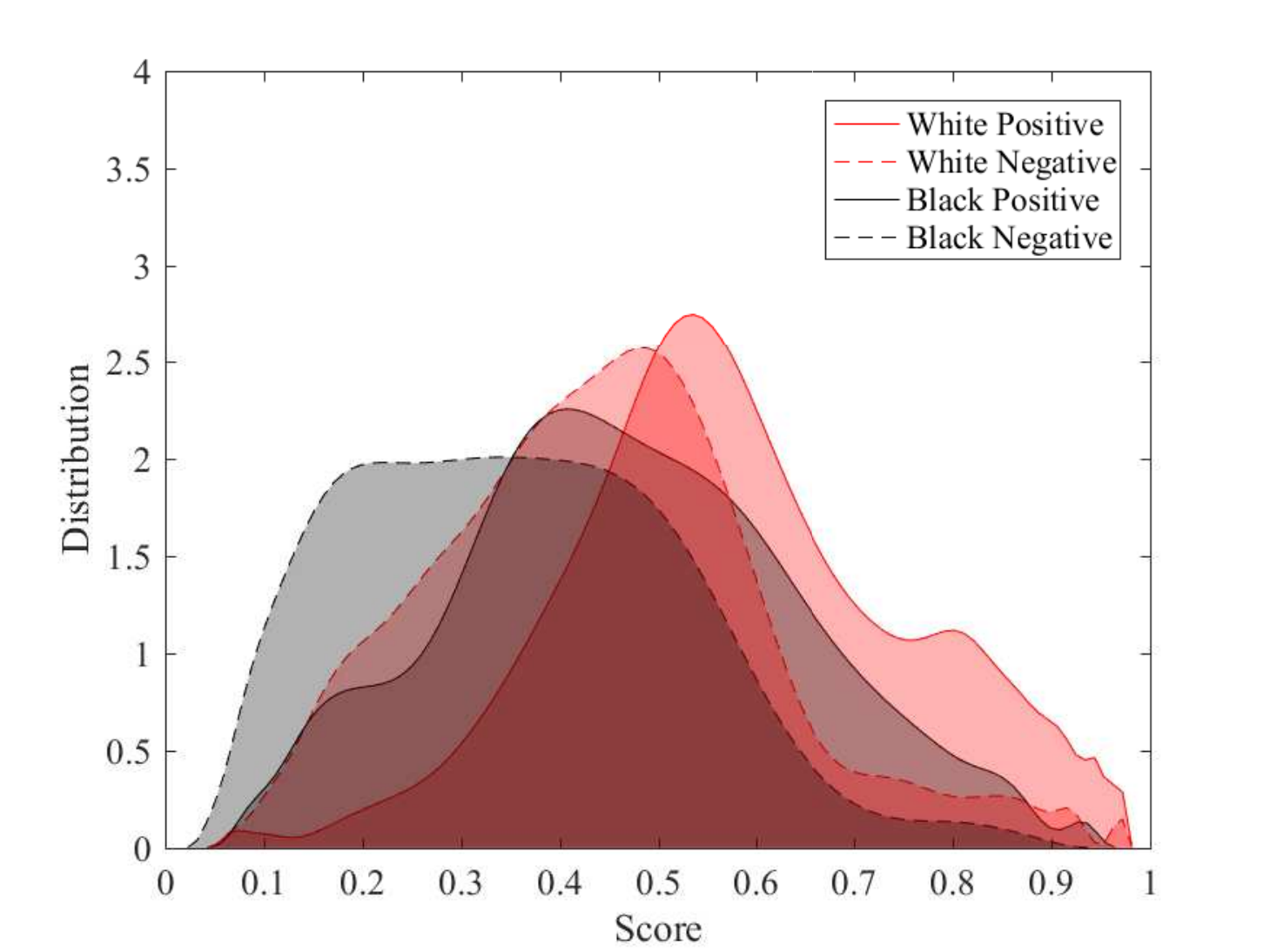}}
      \centerline{(d)}
\end{minipage}
%\end{tabular}
\caption{The score distributions of the white and the black in test set with different $\alpha$, (a) $\alpha=0.01$, (b) $\alpha=0.1$, (c) $\alpha=0.2$, and (d) $\alpha=1$.}
\label{fig:distr}
\end{figure}

We conducted the experiments with different values of the hyperparameter $\alpha$ to explore the influence of $\alpha$ on the performance of the classifier, which is shown in Fig.~\ref{fig:alpha}. When $\alpha$ is $1$, the classifier is the normal classifier with minimum error. When $\alpha$ is close to $0$, the classifier is a mapping to equalize the score distributions in two groups, emphasizing the strong classification parity. When $\alpha$ increases from 0 to 1, weighing more on the accuracy loss, the accuracy of the classifier first increases significantly and then saturates to an upper bound after $\alpha=0.2$. As for the classification parity, the differences of TPR/FPR between two groups also enlarge as $\alpha$ increases, indicating the deteriorate the fairness in ML classifier. Besides, the score distributions of the white and the black in test set with different $\alpha$ values are presented in Figure~\ref{fig:distr}. We can see that the smaller $\alpha$, the closer the score distributions of the positive/negative samples in ``white'' and ``black'' groups. Balancing the accuracy and fairness, we recommend that the best range of $\alpha$ is $0.1-0.2$.

From the experiments, we can see that the proposed training method can provide a classifier equalizing the distribution of the output scores among the groups. Ideally, the classifier has equal and indistinguishable learning performance statistics, e.g., TPR and FPR, among all the protected attributes.

\newpage

\section{Acknowledgements}
This work was initiated at the Workshop for Women in Math and Public Policy.
We would like to thank
IPAM and The Luskin Center
for hosting the workshop
as well as 
the organizers
Mary Lee and
Aisha Najera Chesler
for their tireless efforts.
We also thank the anonymous reviewers for their
constructive comments,
which helped
to significantly
improve the presentation.

\bibliographystyle{plain}
\bibliography{refs}

%%===========================================
%% Appendix:  Bayesian decision/classification notes from Min's group.
%% 
%%  TO do:  copy or move relevant content to the main body.
%%

%\newpage
%\input Appendix_wm.tex

\end{document}